\documentclass[lettersize,journal]{IEEEtran}
\usepackage{amsmath,amsfonts}
\usepackage{array}

\usepackage[caption=false, font=footnotesize, labelformat=empty]{subfig}
\usepackage{textcomp}
\usepackage{stfloats}
\usepackage{url}
\usepackage{verbatim}
\usepackage{graphicx}
\usepackage{cite}
\usepackage{amssymb} 

\usepackage[table]{xcolor}
\definecolor{Viri1Base}{HTML}{440154}
\definecolor{Viri2Base}{HTML}{3B528B}
\definecolor{Viri3Base}{HTML}{21918C}
\definecolor{Viri4Base}{HTML}{5EC962}
\definecolor{Viri5Base}{HTML}{FDE725}

\colorlet{colour1}{Viri1Base!25!white}
\colorlet{colour2}{Viri2Base!25!white}
\colorlet{colour3}{Viri3Base!25!white}
\colorlet{colour4}{Viri4Base!25!white}
\colorlet{colour5}{Viri5Base!25!white}

\usepackage{stackengine}

\usepackage[ruled,vlined]{algorithm2e}

\usepackage{stmaryrd}

\newcolumntype{C}[1]{>{\centering\arraybackslash}m{#1}}
\newcolumntype{R}[1]{>{\raggedright\arraybackslash}m{#1}}
\newcolumntype{L}[1]{>{\raggedleft\arraybackslash}m{#1}}

\makeatletter
\def\bstctlcite{\@ifnextchar[{\@bstctlcite}{\@bstctlcite[@auxout]}}
\def\@bstctlcite[#1]#2{\@bsphack
  \@for\@citeb:=#2\do{%
    \edef\@citeb{\expandafter\@firstofone\@citeb}%
    \if@filesw\immediate\write\csname #1\endcsname{\string\citation{\@citeb}}\fi}%
  \@esphack}
\makeatother

\DeclareMathOperator*{\argmin}{arg\,min}
\DeclareMathOperator*{\argmax}{arg\,max}

\usepackage{booktabs}
\usepackage{multirow}

\usepackage[T1]{fontenc}

\usepackage{calc} 

\usepackage[pdfstartview=XYZ,
bookmarks=true,
colorlinks=true,
linkcolor=blue,
urlcolor=blue,
citecolor=blue,
bookmarks=true,
linktocpage=true, 
hyperindex=true
]{hyperref}

\usepackage{orcidlink}

\begin{document}
\bstctlcite{IEEEexample:BSTcontrol} 

\title{Role-Aware Artificial Intelligence\\Across Augmentation and Automation\\in Human-Machine Symbiosis}

\author{Ching-Chun Chang, Yuchen Guo, Hanrui Wang, Timo Spinde, and Isao Echizen

\thanks{C.-C. Chang, Y. Guo, H. Wang, T. Spinde and I. Echizen are with the Information and Society Research Division, National Institute of Informatics, Tokyo, Japan. Y. Guo and I. Echizen are also with the Graduate School of Information Science and Technology, University of Tokyo, Tokyo, Japan.
}
\thanks{Correspondence: C.-C. Chang (email: ccchang@nii.ac.jp)
}
}

\maketitle

\begin{abstract}
The evolution of artificial intelligence (AI) has rendered the boundary between humanity and computational machinery increasingly ambiguous. In the presence of more interwoven relationships within human-machine symbiosis, the very notion of AI-generated information becomes difficult to define, as such information arises not from either humans or machines in isolation, but from their mutual shaping. At times AI acts in place of the human, automating the task; at others it extends what the human can do, augmenting their capability. Therefore, a more pertinent question lies not merely in whether AI has participated, but in how it has participated. In general, the role assumed by AI is often specified, either implicitly or explicitly, in the input prompt, yet becomes less apparent or altogether unobservable when the generated content alone is available. Once detached from the dialogue context, the functional role may no longer be traceable. This study considers the problem of tracing the functional role played by AI in natural language generation. A methodology is proposed to infer the latent role specified by the prompt, embed this role into the content during the probabilistic generation process and subsequently recover the nature of AI participation from the resulting text. Experimentation is conducted under a representative scenario in which AI acts either as an assistive agent that edits human-written content or as a creative agent that generates new content from a brief concept. The experimental results support the validity of the proposed methodology in terms of discrimination between roles, robustness against perturbations and preservation of linguistic quality. We envision that this study may contribute to future research on the ethics of AI with regard to whether AI has been used fairly, transparently and appropriately.
\end{abstract}

\section{Introduction}
\IEEEPARstart{A}{rtificial} intelligence (AI) has long been shaped by the pursuit of imitating the human mind through computational machinery whose behaviour evokes the belief that a thinking mind resides within it~\cite{Turing:1950aa}. As we devote ever greater energies of our lives to the development of mechanical life, it may be envisaged that human beings and intelligent machines are evolving towards an increasingly intertwined relationship resembling \emph{symbiosis}, in which each compensates for the limitations and extends the capabilities of the other~\cite{butler1863darwin, 4503259, 10.5555/3061053.3061219, Fugener:2025aa}. Whether a machine thinks remains a profound question in its own right; yet human–machine symbiosis brings into view a companion question of how the machine takes part. It is not a question of mere presence or absence, but of the role it plays in relation to humans. This role ranges from a machine that acts on our behalf, taking a task off our hands, to one through which we act, extending what we are able to do. The former is \emph{automation}~\cite{Bainbridge:1983aa}; the latter, \emph{augmentation}~\cite{Engelbart1962Augmenting}.

\begin{figure}[!t]
\centering
\includegraphics[width=1.0\linewidth]{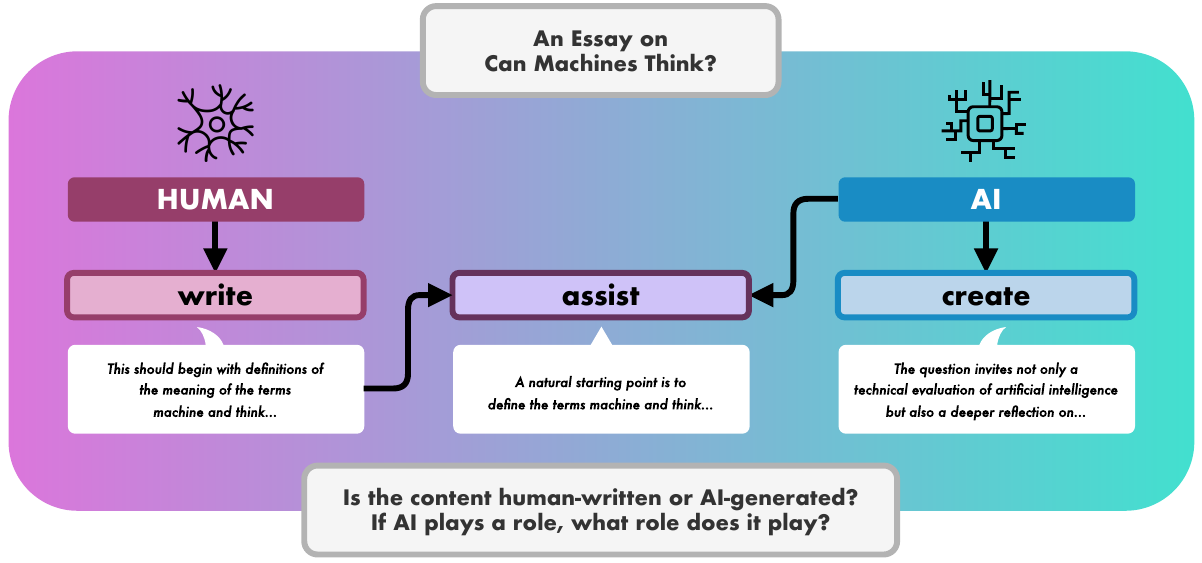}
\caption{An essay on a topic may be written by a human or generated with the involvement of AI, whose role may range from augmentation, where it assists the human author, to automation, where it generates the content more independently; yet the form of human–machine collaboration that occurred may no longer be transparent from the observed essay.}
\label{fig:teaser}
\end{figure}

As AI systems become increasingly involved in human activities, their growing capabilities are accompanied by a range of risks and concerns~\cite{Floridi:2004aa, Russell:2015aa, BrundageEtAl2018, 10.1145/3531146.3533088}. AI models may hallucinate, fabricating factual assertions that are false and misleading~\cite{zhang-etal-2025-sirens}; may elicit toxic content that breaks alignment with ethical principles and social norms~\cite{hendrycks2021aligning}; may reproduce algorithmic bias, reinforcing stereotypical and unfair associations across demographics such as age, gender, race and cultural ethnicity~\cite{Caliskan:2017aa}; and may generate synthetic content used to impersonate individuals, fabricate events and forge evidence~\cite{Chesney:2019aa}.

In response, a substantial body of work has sought to detect AI-generated information~\cite{NEURIPS2019_3e9f0fc9, 9115874, Dathathri:2024ab}. In human-machine collaboration, however, the very notion of AI-generated information becomes difficult to define, for such information emerges not from either humans or machines in isolation, but from their mutual shaping. Therefore, a more pertinent question may not be merely whether AI was involved in producing a piece of information, but how it participated in its production. A passage labelled AI-generated may lie anywhere from text principally authored by a human and merely polished by a model, to text produced by a model almost entirely on its own, as illustrated in Figure~\ref{fig:teaser}. These positions are not equal in what they signify or in how far they may be trusted.

In this study, we consider the problem of tracing the functional role assumed by AI in human-machine collaboration, with a focus on natural language generation. Language is the primary medium of human–machine interaction, as most interaction happens through text, where both the task-specifying prompt and the returned response take the form of language. With large language models increasingly serving as the foundation of AI agents, language also provides an interface through which tasks are specified, plans are formulated, and tools are invoked. We restrict our scope to two contrasting roles: automation, in which AI creates new content from a brief specification, and augmentation, in which AI assists a human author by editing existing content. Such roles may leave little, if any, discernible evidence in the generated textual content. In general, the role assumed by AI is often specified, either implicitly or explicitly, in the input prompt, yet becomes less apparent or altogether unobservable in the resulting output. In other words, the role may become obscured once detached from its dialogue context. This observation motivates our methodology, which analyses the role specified in the prompt and embeds it within the response at the time of generation. The role can later be extracted for resolving the nature of AI participation. The main contributions of this study are summarised as follows:
\begin{itemize}
	\item Problem formulation: We formulate AI traceability beyond binary human-versus-machine detection by considering the functional role played by AI in human–machine collaboration.
	\item Role-aware framework: We propose a framework that infers the latent role from the input prompt and proactively embeds role-specific statistical traces into the generation process, allowing the role to be recovered from the generated text alone, without access to the original prompt or dialogue context.
	\item Experimental validation: We evaluate the system across multiple datasets and language models, examining discriminability, generative quality, robustness, and sensitivity to system hyperparameters.
\end{itemize}

\section{Background}\label{sec:background}
The following preliminaries provide the foundation for this study. We begin by describing the typical generative process of language models and then formulate the problem of inferring the role played by AI from the observed content. We subsequently review prior work on AI content detection in order to contextualise and position the present research.

\subsection{Language Generation}
Contemporary language models typically generate text by repeatedly predicting the next token in a sequence, proceeding one token at a time in an autoregressive loop~\cite{NIPS2017_3f5ee243, NEURIPS2020_1457c0d6, 10.5555/3600270.3602281, 10.5555/3692070.3693470}. Given an input prompt $\mathbf{x}$, the model $f$ autoregressively generates an output sequence $\mathbf{y}$ by iteratively predicting the next token conditioned on the prompt and all previously generated tokens. For notational simplicity, we assume that each text is represented as a sequence of tokens, as language models operate on tokens as the atomic units of text. The tokenisation and detokenisation procedures are therefore omitted throughout this paper. At each generation step $t$, the language model produces a logit vector
\begin{equation}
	\mathbf{z} = f(\mathbf{x}, \mathbf{y}_{<t}) \in \mathbb{R}^{\|\mathcal{V}\|},
\end{equation}
where $\mathcal{V}$ denotes the vocabulary, namely the set of all possible tokens associated with a language model. Applying the softmax function, the logits are transformed into a probability distribution over the next token
\begin{equation}
\mathbb{P}(y_t = v \mid \mathbf{x}, \mathbf{y}_{<t}) = \operatorname{softmax}(\mathbf{z})_v	= \frac{\exp(z_v)}{\sum_{u\in \mathcal{V}} \exp(z_u)} .
\end{equation}
The next token is then sampled according to
\begin{equation}
y_t \sim \operatorname{softmax}(\mathbf{z}).
\end{equation}
This process repeats until the end-of-sequence (EOS) token is generated.

\begin{figure*}[!t]
\centering
\includegraphics[width=1.0\linewidth]{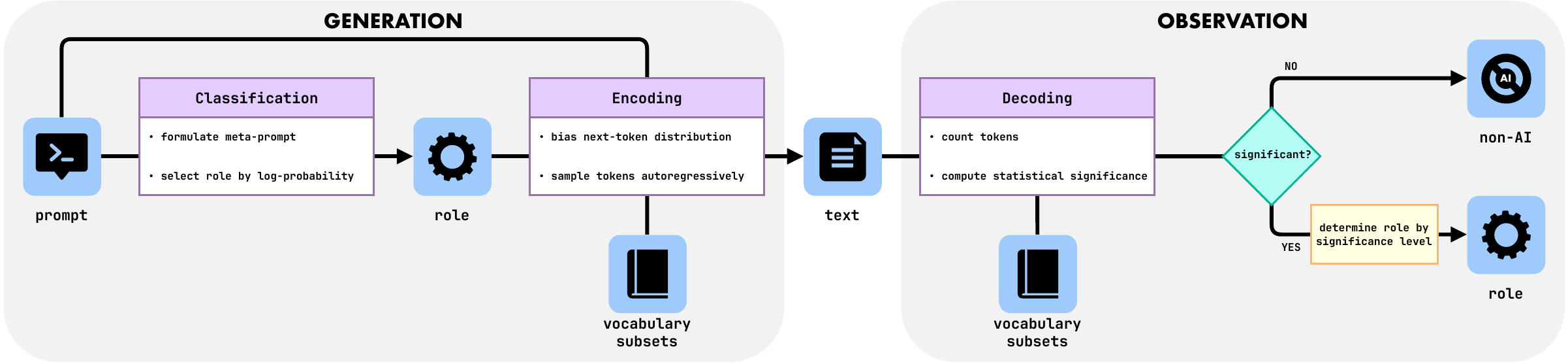}
\caption{Overview of the role-aware AI framework. During generation, the input prompt is classified to infer the latent AI role, which is then encoded into the generated text by biasing the next-token distribution according to role-specific vocabulary subsets. During observation, the role is decoded from the text alone by counting tokens from these subsets and evaluating their statistical significance. If no role yields significant evidence, no AI is considered involved; otherwise, the role with the strongest statistical significance is determined.}
\label{fig:workflow}
\end{figure*}

\subsection{Problem Formulation}
The prompt $\mathbf{x}$ may implicitly or explicitly specify a functional role under which the model is expected to operate. The research problem is to infer, from the observable output $\mathbf{y}$, the underlying role specified by the input $\mathbf{x}$. Let $\mathbf{r} \in \mathcal{R}$ denote a role (expressed as a sequence of tokens), where $\mathcal{R}$ is a predefined role space. Formally, the most plausible latent role $\mathbf{r}^*$ induced by the input prompt $\mathbf{x}$ is defined as
\begin{equation}
\mathbf{r}^* = \argmax_{\mathbf{r} \in \mathcal{R}} \mathbb{P}(\mathbf{r} \mid \mathbf{x}) .
\end{equation}
In practice, however, the source prompt is often unavailable, and only the derived content $\mathbf{y}$ can be observed. The objective is therefore to infer the underlying role from the content alone, that is,
\begin{equation}
\hat{\mathbf{r}} = \argmax_{\mathbf{r} \in \mathcal{R}} \mathbb{P}(\mathbf{r} \mid \mathbf{y}) ,
\end{equation}
where $\hat{\mathbf{r}}$ is expected to coincide with the true role $\mathbf{r}^*$. A role space may include an assistive role, which edits existing content, and a creative role, which generates new content from a concept, together with other role definitions depending on the application context. We define $\hat{r} = \varnothing$ when the language model does not play a role in producing the observed content, in which case the content is regarded as human-written.

\subsection{AI Content Detection}
Prior work on AI content detection can broadly be divided into reactive and proactive paradigms~\cite{11185354}. The reactive paradigm attempts to uncover, after release, identifiable signs left by language models within the content. By contrast, the proactive paradigm attempts to insert, before release, verifiable marks within content generated by language models, such that its origin may later be authenticated. 

For the reactive paradigm, existing methods are typically based on either statistical or data-driven models. Statistical models rely on the uncertainty of a language model with respect to the observed content, since text generation is founded upon the probabilistic sampling of tokens~\cite{10.5555/3053718.3053722, gehrmann-etal-2019-gltr, su-etal-2023-detectllm, pmlr-v202-mitchell23a}. Detection methods based on statistical models are often interpretable and require little, if any, additional training cost. However, such methods may fail when the identifiable signs reside not in observable statistical patterns, but instead in deeper latent representations. Data-driven models learn discriminative features directly from large corpora of human-written and machine-generated texts~\cite{uchendu-etal-2021-turingbench-benchmark}. While such high-dimensional features may be more effective in capturing subtle artefacts left by language models, they are often less interpretable and require substantial labelled data and training effort. In addition, detection methods based on data-driven models may fail to generalise to unseen content that differs from the examples encountered during training.

The proactive paradigm is represented primarily by watermarking methods~\cite{pmlr-v202-kirchenbauer23a}. Rather than searching for discriminative patterns, these methods explicitly embed a mark into the content during the generation process and detection is subsequently performed by verifying its presence or absence. Although watermarking methods may provide greater reliability, they depend upon the premise that the party operating the language model is cooperative and subject to regulation. Consequently, they are not applicable to arbitrary or already-existing content originating from an uncooperative source.

The problem of AI content detection has been studied extensively in prior work, in which the task is typically framed as a binary classification between human-written content and machine-generated content. However, in the presence of increasingly interwoven relationships within human-machine collaboration, the question of how AI has participated in the generation process may extend far beyond such a binary classification. This study seeks to trace the role played by AI from the observed content. The proposed method is founded upon the premise of a cooperative generating party. In particular, it follows the proactive paradigm and is developed on the basis of watermarking techniques.

\section{Methodology}\label{sec:method}
The methodology of this study consists of the following procedures: role classification, role encoding and role decoding. To begin with, the latent role implied by the input prompt is inferred through meta-prompting. The identified role is then encoded within the probabilistic generation process, leaving statistical traces in the generated content, from which the role can subsequently be decoded. An overview of the proposed methodology is illustrated in Figure~\ref{fig:workflow}, while the complete procedure is summarised in Algorithm~\ref{alg:system}.

\subsection{Role Classification}
To operationalise role classification, the input prompt is reformulated into a meta-prompt that treats the original instruction as the object of inference and asks the model to identify the latent role it implies. An example meta-prompt, which defines the candidate roles and asks the model to identify the appropriate one, is given below:
\begin{quote}
	\textit{Definitions:}\\
	\textit{Assistive ---}
	\textit{edit given content.}\\
	\textit{Creative ---}
	\textit{generate content about given concept.}\\
	\textit{Classify this instruction:} $[\mathbf{x}]$
\end{quote}
In-context examples may optionally be incorporated into the meta-prompt to further illustrate how different prompts correspond to distinct roles. Given the meta-prompt $\mathbf{x}_{\Omega}$, the latent role is inferred by selecting the candidate role with the highest length-normalised log-probability
\begin{equation}
\begin{split}
\mathbf{r}^* &= \argmax_{\mathbf{r} \in \mathcal{R} } \frac{1}{\| \mathbf{r} \|} \log \mathbb{P}(\mathbf{r} \mid \mathbf{x}_{\Omega})\\
&= \argmax_{\mathbf{r} \in \mathcal{R} } \frac{1}{\| \mathbf{r} \|} \sum_{t=1}^{\| \mathbf{r} \|} \log \mathbb{P}(r_t \mid \mathbf{x}_{\Omega}, \mathbf{r}_{<t})
\end{split}
\end{equation}
Since the candidate roles may consist of different numbers of tokens, length normalisation is applied to avoid favouring shorter role descriptions.

\subsection{Role Encoding}
Role encoding introduces statistical traces associated with the latent role $\mathbf{r}^*$ into the probabilistic generation process, such that the generated content $\mathbf{y}$ carries evidence of the underlying role. To this end, each candidate role $\mathbf{r} \in \mathcal{R}$ is associated with a designated subset of the vocabulary, denoted by $\mathcal{W}_{\mathbf{r}} \subset \mathcal{V}$. During generation, the language model is biased towards tokens corresponding to $\mathbf{r}^*$, thereby leaving role-specific traces in the generated sequence. Each role-specific token subset is randomly sampled from the vocabulary with equal cardinality
\begin{equation}
\| \mathcal{W}_{\mathbf{r}} \| = q \| \mathcal{V} \| ,
\end{equation}
where $q \in (0, 1)$ denotes the vocabulary coverage rate, controlling the proportion of vocabulary tokens assigned to each role. At generation step $t$, the language model produces a logit vector $\mathbf{z} \in \mathbb{R}^{\|\mathcal{V}\|}$. 
To increase the likelihood of generating tokens associated with $\mathbf{r}^*$, the logits corresponding to the token subset $\mathcal{W}_{\mathbf{r}^*}$ are increased by a constant bias $\delta$, yielding
\begin{equation}
    z'_v =    
    \begin{cases}
        z_v + \delta, & \text{if } v \in \mathcal{W}_{\mathbf{r}^*}, \\
        z_v, & \text{otherwise} .
    \end{cases}
\end{equation}
Consequently, tokens associated with $\mathbf{r}^*$ become more likely to be sampled. Applying the softmax function, the biased logits are converted into a probability distribution
\begin{equation}
\mathbb{P}(y_t = v \mid \mathbf{x}, \mathbf{y}_{<t}) = \operatorname{softmax}(\mathbf{z}')_v = \frac{\exp(z'_v)}{\sum_{u\in \mathcal{V}} \exp(z'_u)}, 
\end{equation}
from which the next token is sampled and appended to the generated sequence. Repeating this procedure throughout generation creates statistical evidence that can later be used to infer the latent role from the generated content alone.

\begin{algorithm}[t!]
\caption{Role-Aware AI}
\label{alg:system}

\vspace{1ex}
\tcp{----- role classification -----}
\SetKwFunction{FClassifier}{Classifier}
\SetKwProg{Fn}{Function}{:}{}
\Fn{\FClassifier{$\mathbf{x}$}}{
formulate meta-prompt $\mathbf{x}_{\Omega}$\\
$\mathbf{r}^* \gets \argmax_{\mathbf{r} \in \mathcal{R} } \frac{1}{\| \mathbf{r} \|} \sum_{t=1}^{\| \mathbf{r} \|} \log \mathbb{P}(r_t \mid \mathbf{x}_{\Omega}, \mathbf{r}_{<t})$\\
\KwRet{$\mathbf{r}^*$}
}

\vspace{1ex}
\tcp{----- role encoding -----}
\SetKwFunction{FEncoder}{Encoder}
\SetKwProg{Fn}{Function}{:}{}
\Fn{\FEncoder{$\mathbf{x}, \mathbf{r}^*$}}{
$t \gets 1$\\
\While{$y_{t-1} \neq \textnormal{EOS}$}{
$\mathbf{z} \gets f(\mathbf{x}, \mathbf{y}_{<t})$\\
	\For{$v \in \mathcal{V}$}{
		\eIf{$v \in \mathcal{W}_{\mathbf{r}^*}$}{
    	$z'_v \gets z_v + \delta$
    	}{$z'_v \gets z_v$}
	}
sample $y_t \sim \operatorname{softmax}(\mathbf{z}')$\\ 
append $y_t$ to $\mathbf{y}$\\
$t \gets t + 1$\\
}
\KwRet{$\mathbf{y}$}
}

\vspace{1ex}
\tcp{----- role decoding -----}
\SetKwFunction{FDecoder}{Decoder}
\SetKwProg{Fn}{Function}{:}{}
\Fn{\FDecoder{$\mathbf{y}$}}{
\For{$\mathbf{r} \in \mathcal{R}$}{
	$n_{\mathbf{r}} \gets \sum_{t=1}^{\| \mathbf{y} \|} \mathbb{I}\bigl( y_t \in \mathcal{W}_{\mathbf{r}} \bigr)$\\
	$p_{\mathbf{r}} \gets \sum_{n=n_{\mathbf{r}}}^{\| \mathbf{y} \|} \binom{\| \mathbf{y} \|}{n} q^{n} (1-q)^{\| \mathbf{y} \|-n}$\\
}
\eIf{$\min_{\mathbf{r} \in \mathcal{R}} p_{\mathbf{r}} \geq \theta$}{
	$\hat{\mathbf{r}} \gets \varnothing$
}{
	$\hat{\mathbf{r}} \gets \argmin_{\mathbf{r} \in \mathcal{R}} p_{\mathbf{r}}$

}
\KwRet{$\hat{\mathbf{r}}$}
}

\vspace{1ex}
\tcp{----- main program -----}
define role space $\mathcal{R}$\\
define vocabulary coverage rate $q$\\
define vocabulary subsets $\mathcal{W}_{\mathbf{r}}$ for all $ \mathbf{r} \in \mathcal{R}$\\
define statistical significance threshold $\theta$\\
$\mathbf{r}^* \gets$ \FClassifier{$\mathbf{x}$}\\
$\mathbf{y} \gets$ \FEncoder{$\mathbf{x}, \mathbf{r}^*$}\\
$\hat{\mathbf{r}} \gets$ \FDecoder{$\mathbf{y}$}\\

\end{algorithm}

\subsection{Role Decoding}
Role decoding seeks to determine the latent role from an observed text sequence $\mathbf{y}$. The decoding procedure evaluates whether $\mathbf{y}$ contains an unusually large number of tokens from any vocabulary subset $\mathcal{W}_{\mathbf{r}}$. Under the null hypothesis $H_0$, the observed sequence is assumed not to exhibit any preference for $\mathcal{W}_{\mathbf{r}}$. Consequently, each generated token $y_t$ independently falls into $\mathcal{W}_{\mathbf{r}}$ with probability $q$, that is,
\begin{equation}
H_0: \mathbb{P} ( y_t \in \mathcal{W}_{\mathbf{r}} ) = q = \frac{\|\mathcal{W}_{\mathbf{r}}\|}{\|\mathcal{V}\|}  .
\end{equation}
Let $N$ denote the random variable representing the number of tokens that belong to $\mathcal{W}_{\mathbf{r}}$ under $H_0$. It then follows that 
\begin{equation}
N \sim \operatorname{Binomial}(\| \mathbf{y} \|, q).
\end{equation}
Its probability mass function is therefore
\begin{equation}
\mathbb{P}(N=n) = \binom{\| \mathbf{y} \|}{n} q^{n} (1-q)^{\| \mathbf{y} \|-n}.
\end{equation}
For an observed generated sequence $\mathbf{y}$, the number of tokens belonging to $\mathcal{W}_{\mathbf{r}}$ is computed as
\begin{equation}
n_{\mathbf{r}} = \sum_{t=1}^{\| \mathbf{y} \|} \mathbb{I}\bigl( y_t \in \mathcal{W}_{\mathbf{r}} \bigr),
\end{equation}
The statistical significance of the role $\mathbf{r}$ is then quantified by the upper-tail probability
\begin{equation}
p_{\mathbf{r}} = \mathbb{P} \bigl(N \geq n_{\mathbf{r}} \bigr)
= \sum_{n=n_{\mathbf{r}}}^{\| \mathbf{y} \|} \binom{\| \mathbf{y} \|}{n} q^{n} (1-q)^{\| \mathbf{y} \|-n}
\end{equation}
A smaller $p_{\mathbf{r}}$ indicates stronger evidence that $\mathbf{y}$ contains more tokens from $\mathcal{W}_{\mathbf{r}}$ than would be expected by chance, and is therefore more likely to have been generated under role $\mathbf{r}$. Finally, the role is determined according to the minimum $p$-value among all candidates. If no candidate yields statistically significant evidence, that is, if the minimum $p$-value exceeds the significance threshold $\theta$, no role is assigned. The resulting decision rule is given by
\begin{equation}
\hat{\mathbf{r}} =    
\begin{cases}
	\varnothing, & \text{if } \min_{\mathbf{r} \in \mathcal{R}} p_{\mathbf{r}} \geq \theta, \\
	\argmin_{\mathbf{r} \in \mathcal{R}} p_{\mathbf{r}}, & \text{otherwise}.
\end{cases}
\end{equation}

\section{Evaluation}\label{sec:eval}
The experimentation in this study focuses on a restricted but well-defined scenario in which AI acts either as an assistive agent that edits human-written content or as a creative agent that generates text about target concepts. The proposed method is evaluated from three complementary perspectives: discriminability, robustness and perplexity.

\subsection{Experimental Setup}
\begin{figure}[!t]
\centering
\includegraphics[width=0.95\linewidth]{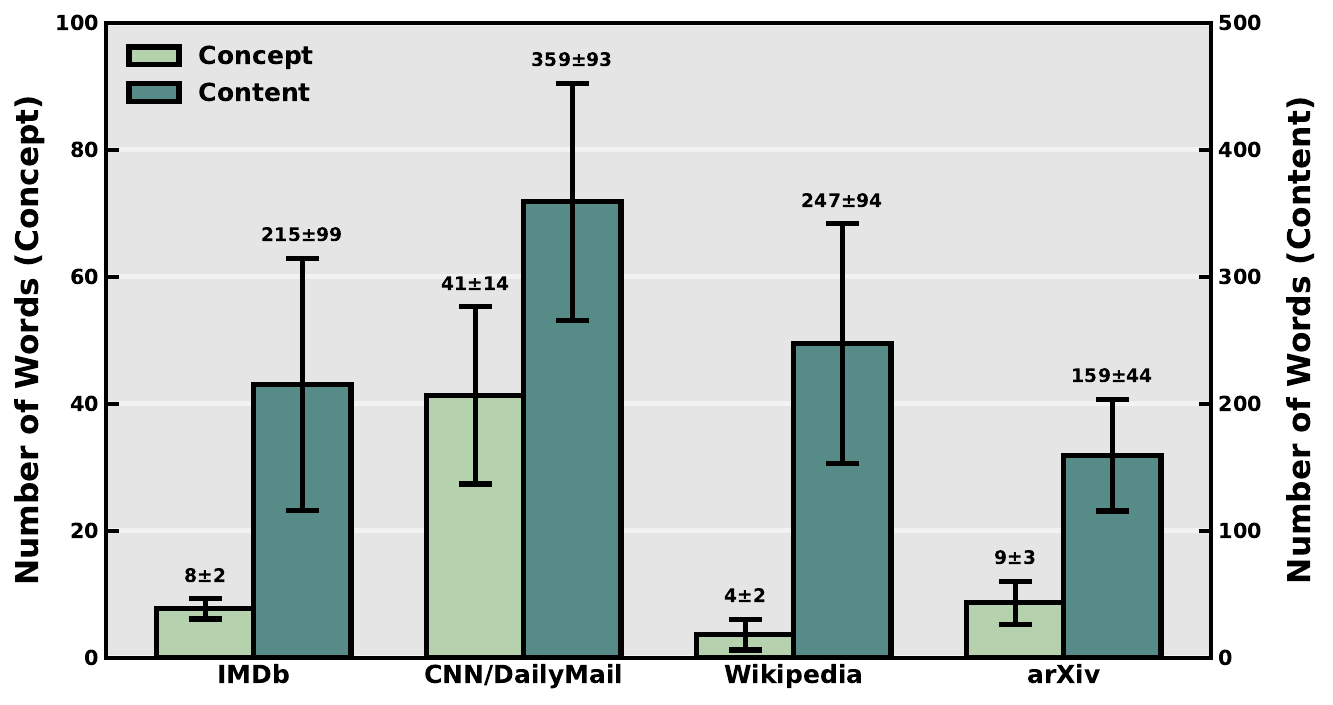}
\caption{Statistics of concept and content lengths across datasets.}
\label{fig:dataset_info}
\end{figure}

\paragraph*{Datasets}
Experiments are conducted on four benchmark datasets spanning diverse domains, including entertainment, journalism, general knowledge, and scientific literature:
\begin{itemize}
	\item IMDb dataset consists of movie reviews annotated with binary sentiment labels (positive or negative)~\cite{maas-etal-2011-learning}. We define `\texttt{<sentiment>} review of the film \texttt{<title>}' as the concept and the corresponding \texttt{<review>} as the content.
	\item CNN/DailyMail dataset consists of journalistic articles paired with summarised highlights~\cite{chen-etal-2016-thorough}. We define the \texttt{<highlight>} as the concept and the corresponding \texttt{<article>} as the content.
	\item Wikipedia dataset consists of encyclopaedic topics accompanied by descriptive summaries~\cite{merity2017pointer}. We define the \texttt{<topic>} as the concept and the corresponding \texttt{<description>} as the content.
	\item arXiv dataset consists of scientific research papers with titles and abstracts~\cite{clement2019usearxivdataset}. We define the \texttt{<title>} as the concept and the corresponding \texttt{<abstract>} as the content.
\end{itemize}
A total of 1,000 evaluation samples are randomly selected from each dataset, subject to the constraint that each concept contains fewer than 100 words and its corresponding content contains 100 to 500 words. The word distribution for each dataset is shown in Figure~\ref{fig:dataset_info}.

\begin{table*}[t]
\centering
\caption{
Pairwise binary classification performance across models and datasets\\
Tiers: \colorbox{colour1}{$\le 0.6$}\colorbox{colour2}{$\le 0.7$}\colorbox{colour3}{$\le 0.8$}\colorbox{colour4}{$\le 0.9$}\colorbox{colour5}{$\le 1.0$}\\
}
\label{tab:results}

\setlength{\tabcolsep}{3pt}
\renewcommand{\arraystretch}{1.60}

\resizebox{\linewidth}{!}{%
\begin{tabular}{c | c | cccccccccccc | cccccccccccc}
\toprule

\multicolumn{1}{c}{} &
\multicolumn{1}{c}{} &
\multicolumn{12}{c}{GPT-2 (124 million parameters)} &
\multicolumn{12}{c}{LLaMA-3-Instruct (3 billion parameters)}\\

\cmidrule(lr){3-14}
\cmidrule(lr){15-26}

\multicolumn{1}{c}{} &
\multicolumn{1}{c}{} &

\multicolumn{2}{c}{Entropy} &
\multicolumn{2}{c}{LogLikelihood} &
\multicolumn{2}{c}{LogRank} &
\multicolumn{2}{c}{Curvature} &
\multicolumn{2}{c}{RoBERTa} &
\multicolumn{2}{c}{Ours} &

\multicolumn{2}{c}{Entropy} &
\multicolumn{2}{c}{LogLikelihood} &
\multicolumn{2}{c}{LogRank} &
\multicolumn{2}{c}{Curvature} &
\multicolumn{2}{c}{RoBERTa} &
\multicolumn{2}{c}{Ours} \\

\cmidrule(lr){3-14}
\cmidrule(lr){15-26}

\multicolumn{1}{c}{} &
\multicolumn{1}{c}{} &

AUC & ACC &
AUC & ACC &
AUC & ACC &
AUC & ACC &
AUC & ACC &
AUC & \multicolumn{1}{c}{ACC} &

AUC & ACC &
AUC & ACC &
AUC & ACC &
AUC & ACC &
AUC & ACC &
AUC & ACC \\

\midrule

\multirow{4}{*}{\rotatebox{90}{IMDb}} & $\mathbb{H}$ vs $\mathbb{A}$ & \cellcolor{colour4}0.82 & \cellcolor{colour3}0.74 & \cellcolor{colour3}0.71 & \cellcolor{colour2}0.64 & \cellcolor{colour5}0.96 & \cellcolor{colour5}0.91 & \cellcolor{colour4}0.89 & \cellcolor{colour4}0.81 & \cellcolor{colour5}0.95 & \cellcolor{colour4}0.90 & \cellcolor{colour5}0.93 & \cellcolor{colour4}0.86 & \cellcolor{colour5}0.97 & \cellcolor{colour5}0.91 & \cellcolor{colour5}0.98 & \cellcolor{colour5}0.93 & \cellcolor{colour5}0.98 & \cellcolor{colour5}0.92 & \cellcolor{colour3}0.79 & \cellcolor{colour3}0.74 & \cellcolor{colour4}0.86 & \cellcolor{colour4}0.81 & \cellcolor{colour5}1.00 & \cellcolor{colour5}0.97\\
 & $\mathbb{H}$ vs $\mathbb{C}$ & \cellcolor{colour5}0.91 & \cellcolor{colour4}0.83 & \cellcolor{colour5}0.95 & \cellcolor{colour4}0.88 & \cellcolor{colour5}1.00 & \cellcolor{colour5}0.97 & \cellcolor{colour5}0.94 & \cellcolor{colour4}0.87 & \cellcolor{colour5}0.96 & \cellcolor{colour5}0.92 & \cellcolor{colour5}1.00 & \cellcolor{colour5}1.00 & \cellcolor{colour5}0.99 & \cellcolor{colour5}0.98 & \cellcolor{colour5}1.00 & \cellcolor{colour5}1.00 & \cellcolor{colour5}1.00 & \cellcolor{colour5}0.99 & \cellcolor{colour5}0.91 & \cellcolor{colour4}0.86 & \cellcolor{colour5}0.98 & \cellcolor{colour5}0.95 & \cellcolor{colour5}0.99 & \cellcolor{colour5}0.95\\
 & $\mathbb{A}$ vs $\mathbb{C}$ & \cellcolor{colour1}0.32 & \cellcolor{colour1}0.51 & \cellcolor{colour4}0.89 & \cellcolor{colour4}0.82 & \cellcolor{colour1}0.13 & \cellcolor{colour1}0.51 & \cellcolor{colour2}0.61 & \cellcolor{colour1}0.58 & \cellcolor{colour1}0.58 & \cellcolor{colour1}0.56 & \cellcolor{colour4}0.90 & \cellcolor{colour4}0.85 & \cellcolor{colour1}0.09 & \cellcolor{colour1}0.50 & \cellcolor{colour5}0.95 & \cellcolor{colour5}0.92 & \cellcolor{colour1}0.04 & \cellcolor{colour1}0.50 & \cellcolor{colour4}0.82 & \cellcolor{colour3}0.78 & \cellcolor{colour3}0.75 & \cellcolor{colour2}0.69 & \cellcolor{colour5}1.00 & \cellcolor{colour5}0.98\\
 & Average & \cellcolor{colour2}0.68 & \cellcolor{colour2}0.70 & \cellcolor{colour4}0.85 & \cellcolor{colour3}0.78 & \cellcolor{colour2}0.69 & \cellcolor{colour3}0.80 & \cellcolor{colour4}0.81 & \cellcolor{colour3}0.75 & \cellcolor{colour4}0.83 & \cellcolor{colour3}0.79 & \cellcolor{colour5}0.94 & \cellcolor{colour4}0.90 & \cellcolor{colour2}0.68 & \cellcolor{colour3}0.80 & \cellcolor{colour5}0.98 & \cellcolor{colour5}0.95 & \cellcolor{colour2}0.67 & \cellcolor{colour3}0.80 & \cellcolor{colour4}0.84 & \cellcolor{colour3}0.79 & \cellcolor{colour4}0.86 & \cellcolor{colour4}0.82 & \cellcolor{colour5}0.99 & \cellcolor{colour5}0.97\\

\midrule

\multirow{4}{*}{\rotatebox{90}{CNN/DailyMail}} & $\mathbb{H}$ vs $\mathbb{A}$ & \cellcolor{colour1}0.41 & \cellcolor{colour1}0.51 & \cellcolor{colour1}0.14 & \cellcolor{colour1}0.51 & \cellcolor{colour2}0.67 & \cellcolor{colour2}0.61 & \cellcolor{colour1}0.39 & \cellcolor{colour1}0.50 & \cellcolor{colour5}0.99 & \cellcolor{colour5}0.96 & \cellcolor{colour5}0.96 & \cellcolor{colour5}0.95 & \cellcolor{colour3}0.75 & \cellcolor{colour2}0.67 & \cellcolor{colour4}0.85 & \cellcolor{colour3}0.77 & \cellcolor{colour4}0.84 & \cellcolor{colour3}0.76 & \cellcolor{colour2}0.63 & \cellcolor{colour2}0.62 & \cellcolor{colour3}0.71 & \cellcolor{colour3}0.71 & \cellcolor{colour5}0.99 & \cellcolor{colour5}0.94\\
 & $\mathbb{H}$ vs $\mathbb{C}$ & \cellcolor{colour1}0.36 & \cellcolor{colour1}0.50 & \cellcolor{colour1}0.45 & \cellcolor{colour1}0.51 & \cellcolor{colour5}0.94 & \cellcolor{colour4}0.87 & \cellcolor{colour1}0.50 & \cellcolor{colour1}0.53 & \cellcolor{colour5}0.99 & \cellcolor{colour5}0.96 & \cellcolor{colour5}0.97 & \cellcolor{colour4}0.90 & \cellcolor{colour2}0.70 & \cellcolor{colour2}0.69 & \cellcolor{colour4}0.86 & \cellcolor{colour4}0.83 & \cellcolor{colour4}0.85 & \cellcolor{colour4}0.83 & \cellcolor{colour1}0.49 & \cellcolor{colour1}0.53 & \cellcolor{colour5}0.96 & \cellcolor{colour5}0.93 & \cellcolor{colour5}0.97 & \cellcolor{colour5}0.92\\
 & $\mathbb{A}$ vs $\mathbb{C}$ & \cellcolor{colour1}0.53 & \cellcolor{colour1}0.53 & \cellcolor{colour4}0.86 & \cellcolor{colour3}0.79 & \cellcolor{colour1}0.15 & \cellcolor{colour1}0.51 & \cellcolor{colour1}0.60 & \cellcolor{colour1}0.57 & \cellcolor{colour1}0.50 & \cellcolor{colour1}0.51 & \cellcolor{colour4}0.86 & \cellcolor{colour4}0.85 & \cellcolor{colour1}0.51 & \cellcolor{colour1}0.56 & \cellcolor{colour2}0.64 & \cellcolor{colour2}0.62 & \cellcolor{colour1}0.35 & \cellcolor{colour1}0.53 & \cellcolor{colour1}0.39 & \cellcolor{colour1}0.52 & \cellcolor{colour3}0.78 & \cellcolor{colour3}0.73 & \cellcolor{colour5}0.99 & \cellcolor{colour5}0.97\\
 & Average & \cellcolor{colour1}0.43 & \cellcolor{colour1}0.52 & \cellcolor{colour1}0.48 & \cellcolor{colour1}0.60 & \cellcolor{colour1}0.59 & \cellcolor{colour2}0.66 & \cellcolor{colour1}0.50 & \cellcolor{colour1}0.53 & \cellcolor{colour4}0.83 & \cellcolor{colour4}0.81 & \cellcolor{colour5}0.93 & \cellcolor{colour4}0.90 & \cellcolor{colour2}0.65 & \cellcolor{colour2}0.64 & \cellcolor{colour3}0.78 & \cellcolor{colour3}0.74 & \cellcolor{colour2}0.68 & \cellcolor{colour2}0.70 & \cellcolor{colour1}0.50 & \cellcolor{colour1}0.56 & \cellcolor{colour4}0.82 & \cellcolor{colour3}0.79 & \cellcolor{colour5}0.98 & \cellcolor{colour5}0.94\\

\midrule

\multirow{4}{*}{\rotatebox{90}{Wikipedia}} & $\mathbb{H}$ vs $\mathbb{A}$ & \cellcolor{colour1}0.55 & \cellcolor{colour1}0.52 & \cellcolor{colour1}0.24 & \cellcolor{colour1}0.50 & \cellcolor{colour4}0.84 & \cellcolor{colour3}0.77 & \cellcolor{colour2}0.62 & \cellcolor{colour1}0.59 & \cellcolor{colour5}0.94 & \cellcolor{colour4}0.88 & \cellcolor{colour5}0.96 & \cellcolor{colour5}0.91 & \cellcolor{colour3}0.77 & \cellcolor{colour2}0.69 & \cellcolor{colour4}0.82 & \cellcolor{colour3}0.74 & \cellcolor{colour3}0.80 & \cellcolor{colour3}0.72 & \cellcolor{colour3}0.71 & \cellcolor{colour2}0.65 & \cellcolor{colour3}0.72 & \cellcolor{colour2}0.70 & \cellcolor{colour5}0.96 & \cellcolor{colour4}0.90\\
 & $\mathbb{H}$ vs $\mathbb{C}$ & \cellcolor{colour1}0.51 & \cellcolor{colour1}0.49 & \cellcolor{colour2}0.68 & \cellcolor{colour2}0.63 & \cellcolor{colour5}0.97 & \cellcolor{colour5}0.92 & \cellcolor{colour3}0.77 & \cellcolor{colour3}0.71 & \cellcolor{colour5}0.95 & \cellcolor{colour4}0.90 & \cellcolor{colour5}1.00 & \cellcolor{colour5}0.97 & \cellcolor{colour5}0.97 & \cellcolor{colour5}0.92 & \cellcolor{colour5}1.00 & \cellcolor{colour5}0.97 & \cellcolor{colour5}0.99 & \cellcolor{colour5}0.96 & \cellcolor{colour4}0.83 & \cellcolor{colour3}0.74 & \cellcolor{colour5}0.96 & \cellcolor{colour5}0.91 & \cellcolor{colour5}0.95 & \cellcolor{colour4}0.89\\
 & $\mathbb{A}$ vs $\mathbb{C}$ & \cellcolor{colour1}0.54 & \cellcolor{colour1}0.53 & \cellcolor{colour4}0.88 & \cellcolor{colour3}0.80 & \cellcolor{colour1}0.13 & \cellcolor{colour1}0.50 & \cellcolor{colour2}0.66 & \cellcolor{colour1}0.60 & \cellcolor{colour1}0.57 & \cellcolor{colour1}0.54 & \cellcolor{colour4}0.90 & \cellcolor{colour4}0.88 & \cellcolor{colour1}0.11 & \cellcolor{colour1}0.50 & \cellcolor{colour5}0.97 & \cellcolor{colour5}0.91 & \cellcolor{colour1}0.04 & \cellcolor{colour1}0.50 & \cellcolor{colour2}0.67 & \cellcolor{colour2}0.63 & \cellcolor{colour4}0.86 & \cellcolor{colour3}0.77 & \cellcolor{colour5}0.99 & \cellcolor{colour5}0.96\\
 & Average & \cellcolor{colour1}0.53 & \cellcolor{colour1}0.51 & \cellcolor{colour1}0.60 & \cellcolor{colour2}0.64 & \cellcolor{colour2}0.65 & \cellcolor{colour3}0.73 & \cellcolor{colour2}0.68 & \cellcolor{colour2}0.63 & \cellcolor{colour4}0.82 & \cellcolor{colour3}0.77 & \cellcolor{colour5}0.95 & \cellcolor{colour5}0.92 & \cellcolor{colour2}0.62 & \cellcolor{colour3}0.71 & \cellcolor{colour5}0.93 & \cellcolor{colour4}0.87 & \cellcolor{colour2}0.61 & \cellcolor{colour3}0.73 & \cellcolor{colour3}0.74 & \cellcolor{colour2}0.67 & \cellcolor{colour4}0.85 & \cellcolor{colour3}0.79 & \cellcolor{colour5}0.97 & \cellcolor{colour5}0.92\\

\midrule

\multirow{4}{*}{\rotatebox{90}{arXiv}} & $\mathbb{H}$ vs $\mathbb{A}$ & \cellcolor{colour4}0.85 & \cellcolor{colour3}0.79 & \cellcolor{colour2}0.64 & \cellcolor{colour2}0.61 & \cellcolor{colour5}0.96 & \cellcolor{colour4}0.90 & \cellcolor{colour3}0.76 & \cellcolor{colour2}0.70 & \cellcolor{colour5}0.92 & \cellcolor{colour4}0.85 & \cellcolor{colour4}0.88 & \cellcolor{colour3}0.79 & \cellcolor{colour4}0.89 & \cellcolor{colour4}0.83 & \cellcolor{colour5}0.91 & \cellcolor{colour4}0.87 & \cellcolor{colour4}0.90 & \cellcolor{colour4}0.85 & \cellcolor{colour4}0.86 & \cellcolor{colour4}0.81 & \cellcolor{colour2}0.65 & \cellcolor{colour2}0.61 & \cellcolor{colour5}1.00 & \cellcolor{colour5}0.96\\
 & $\mathbb{H}$ vs $\mathbb{C}$ & \cellcolor{colour4}0.87 & \cellcolor{colour4}0.81 & \cellcolor{colour4}0.89 & \cellcolor{colour3}0.80 & \cellcolor{colour5}0.99 & \cellcolor{colour5}0.96 & \cellcolor{colour4}0.87 & \cellcolor{colour3}0.78 & \cellcolor{colour5}0.92 & \cellcolor{colour4}0.85 & \cellcolor{colour5}0.99 & \cellcolor{colour5}0.96 & \cellcolor{colour5}1.00 & \cellcolor{colour5}0.98 & \cellcolor{colour5}1.00 & \cellcolor{colour5}0.99 & \cellcolor{colour5}1.00 & \cellcolor{colour5}0.99 & \cellcolor{colour5}0.99 & \cellcolor{colour5}0.95 & \cellcolor{colour4}0.90 & \cellcolor{colour4}0.82 & \cellcolor{colour5}1.00 & \cellcolor{colour5}0.97\\
 & $\mathbb{A}$ vs $\mathbb{C}$ & \cellcolor{colour1}0.46 & \cellcolor{colour1}0.49 & \cellcolor{colour4}0.81 & \cellcolor{colour3}0.73 & \cellcolor{colour1}0.22 & \cellcolor{colour1}0.50 & \cellcolor{colour2}0.66 & \cellcolor{colour1}0.59 & \cellcolor{colour1}0.51 & \cellcolor{colour1}0.49 & \cellcolor{colour2}0.64 & \cellcolor{colour1}0.60 & \cellcolor{colour1}0.12 & \cellcolor{colour1}0.50 & \cellcolor{colour5}0.98 & \cellcolor{colour5}0.95 & \cellcolor{colour1}0.02 & \cellcolor{colour1}0.50 & \cellcolor{colour4}0.89 & \cellcolor{colour3}0.80 & \cellcolor{colour4}0.81 & \cellcolor{colour3}0.74 & \cellcolor{colour5}1.00 & \cellcolor{colour5}0.99\\
 & Average & \cellcolor{colour3}0.73 & \cellcolor{colour2}0.70 & \cellcolor{colour3}0.78 & \cellcolor{colour3}0.71 & \cellcolor{colour3}0.72 & \cellcolor{colour3}0.79 & \cellcolor{colour3}0.76 & \cellcolor{colour2}0.69 & \cellcolor{colour3}0.78 & \cellcolor{colour3}0.73 & \cellcolor{colour4}0.83 & \cellcolor{colour3}0.79 & \cellcolor{colour2}0.67 & \cellcolor{colour3}0.77 & \cellcolor{colour5}0.96 & \cellcolor{colour5}0.94 & \cellcolor{colour2}0.64 & \cellcolor{colour3}0.78 & \cellcolor{colour5}0.91 & \cellcolor{colour4}0.85 & \cellcolor{colour3}0.79 & \cellcolor{colour3}0.72 & \cellcolor{colour5}1.00 & \cellcolor{colour5}0.98\\

\midrule

\multirow{4}{*}{\rotatebox{90}{All Datasets}} & $\mathbb{H}$ vs $\mathbb{A}$ & \cellcolor{colour2}0.65 & \cellcolor{colour2}0.64 & \cellcolor{colour1}0.43 & \cellcolor{colour1}0.57 & \cellcolor{colour4}0.86 & \cellcolor{colour3}0.80 & \cellcolor{colour2}0.66 & \cellcolor{colour2}0.65 & \cellcolor{colour5}0.95 & \cellcolor{colour4}0.90 & \cellcolor{colour5}0.93 & \cellcolor{colour4}0.88 & \cellcolor{colour4}0.84 & \cellcolor{colour3}0.78 & \cellcolor{colour4}0.89 & \cellcolor{colour4}0.83 & \cellcolor{colour4}0.88 & \cellcolor{colour4}0.81 & \cellcolor{colour3}0.75 & \cellcolor{colour2}0.70 & \cellcolor{colour3}0.73 & \cellcolor{colour3}0.71 & \cellcolor{colour5}0.98 & \cellcolor{colour5}0.94\\
 & $\mathbb{H}$ vs $\mathbb{C}$ & \cellcolor{colour2}0.66 & \cellcolor{colour2}0.66 & \cellcolor{colour3}0.74 & \cellcolor{colour3}0.71 & \cellcolor{colour5}0.97 & \cellcolor{colour5}0.93 & \cellcolor{colour3}0.77 & \cellcolor{colour3}0.72 & \cellcolor{colour5}0.95 & \cellcolor{colour5}0.91 & \cellcolor{colour5}0.99 & \cellcolor{colour5}0.96 & \cellcolor{colour5}0.92 & \cellcolor{colour4}0.89 & \cellcolor{colour5}0.96 & \cellcolor{colour5}0.95 & \cellcolor{colour5}0.96 & \cellcolor{colour5}0.94 & \cellcolor{colour3}0.80 & \cellcolor{colour3}0.77 & \cellcolor{colour5}0.95 & \cellcolor{colour4}0.90 & \cellcolor{colour5}0.98 & \cellcolor{colour5}0.93\\
 & $\mathbb{A}$ vs $\mathbb{C}$ & \cellcolor{colour1}0.46 & \cellcolor{colour1}0.51 & \cellcolor{colour4}0.86 & \cellcolor{colour3}0.78 & \cellcolor{colour1}0.16 & \cellcolor{colour1}0.50 & \cellcolor{colour2}0.63 & \cellcolor{colour1}0.59 & \cellcolor{colour1}0.54 & \cellcolor{colour1}0.52 & \cellcolor{colour4}0.83 & \cellcolor{colour3}0.80 & \cellcolor{colour1}0.21 & \cellcolor{colour1}0.51 & \cellcolor{colour4}0.89 & \cellcolor{colour4}0.85 & \cellcolor{colour1}0.11 & \cellcolor{colour1}0.51 & \cellcolor{colour2}0.69 & \cellcolor{colour2}0.68 & \cellcolor{colour3}0.80 & \cellcolor{colour3}0.73 & \cellcolor{colour5}0.99 & \cellcolor{colour5}0.98\\
 & Average & \cellcolor{colour1}0.59 & \cellcolor{colour2}0.61 & \cellcolor{colour2}0.68 & \cellcolor{colour2}0.69 & \cellcolor{colour2}0.66 & \cellcolor{colour3}0.74 & \cellcolor{colour2}0.69 & \cellcolor{colour2}0.65 & \cellcolor{colour4}0.81 & \cellcolor{colour3}0.78 & \cellcolor{colour5}0.91 & \cellcolor{colour4}0.88 & \cellcolor{colour2}0.66 & \cellcolor{colour3}0.73 & \cellcolor{colour5}0.91 & \cellcolor{colour4}0.88 & \cellcolor{colour2}0.65 & \cellcolor{colour3}0.75 & \cellcolor{colour3}0.75 & \cellcolor{colour3}0.72 & \cellcolor{colour4}0.83 & \cellcolor{colour3}0.78 & \cellcolor{colour5}0.99 & \cellcolor{colour5}0.95\\

\bottomrule

\end{tabular}
}
\end{table*}

\paragraph*{Models}
For natural language generation, two open-source models are employed:
\begin{itemize}
	\item GPT-2 (base version with 124 million parameters) developed by OpenAI.
	\item LLaMA-3 (instruction-tuned version with 3 billion parameters) developed by Meta.
\end{itemize}
Using the evaluation samples from each dataset, each model generates 1,000 texts acting as an assistive agent conditioned on the human-written content and 1,000 texts acting as a creative agent conditioned on the human-written concepts. For assistive writing, the prompts follow the template:
\begin{equation*}
	\text{ Please } \texttt{<verb>} \text{ the following paragraph: } \texttt{<content>} ,
\end{equation*}
where \texttt{<verb>} is drawn at random from edit, improve, proofread, revise and summarise. For creative writing, the prompts are instead constructed according to the form:
\begin{equation*}
	\text{ Please } \texttt{<verb>} \text{ a paragraph about: } \texttt{<concept>} ,
\end{equation*}
where \texttt{<verb>} is chosen randomly from compose, create, draft, generate and write.

\paragraph*{Hyperparameters}
This study focuses on the scenario in which AI acts as either an assistive or a creative agent; accordingly, the role space $\mathcal{R}$ is defined as $\{\text{assistive}, \text{creative}\}$. The vocabulary coverage rate $q$ is set to 0.5, such that half of the vocabulary is assigned higher sampling probability. The bias strength $\delta$ is set to 1.5 for GPT-2 and 3.0 for LLaMA-3-Instruct. The significance threshold $\theta$ is selected adaptively via multi-fold cross-validation.

\begin{figure}[t!]
    \centering
    \subfloat[GPT-2]{
    	\includegraphics[width=0.49\columnwidth]{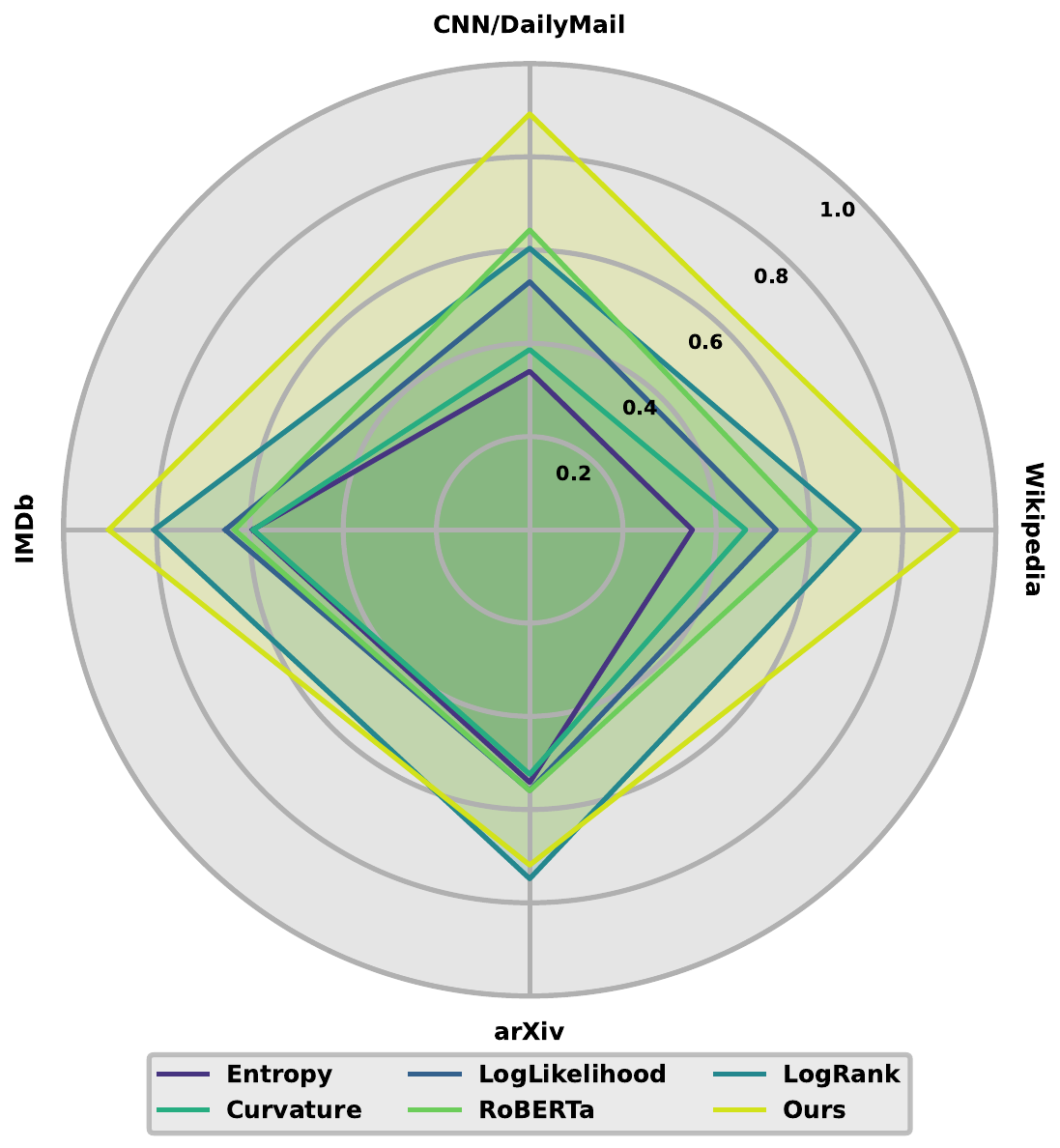}
    }
    \subfloat[LLaMA-3-Instruct ]{
        \includegraphics[width=0.49\columnwidth]{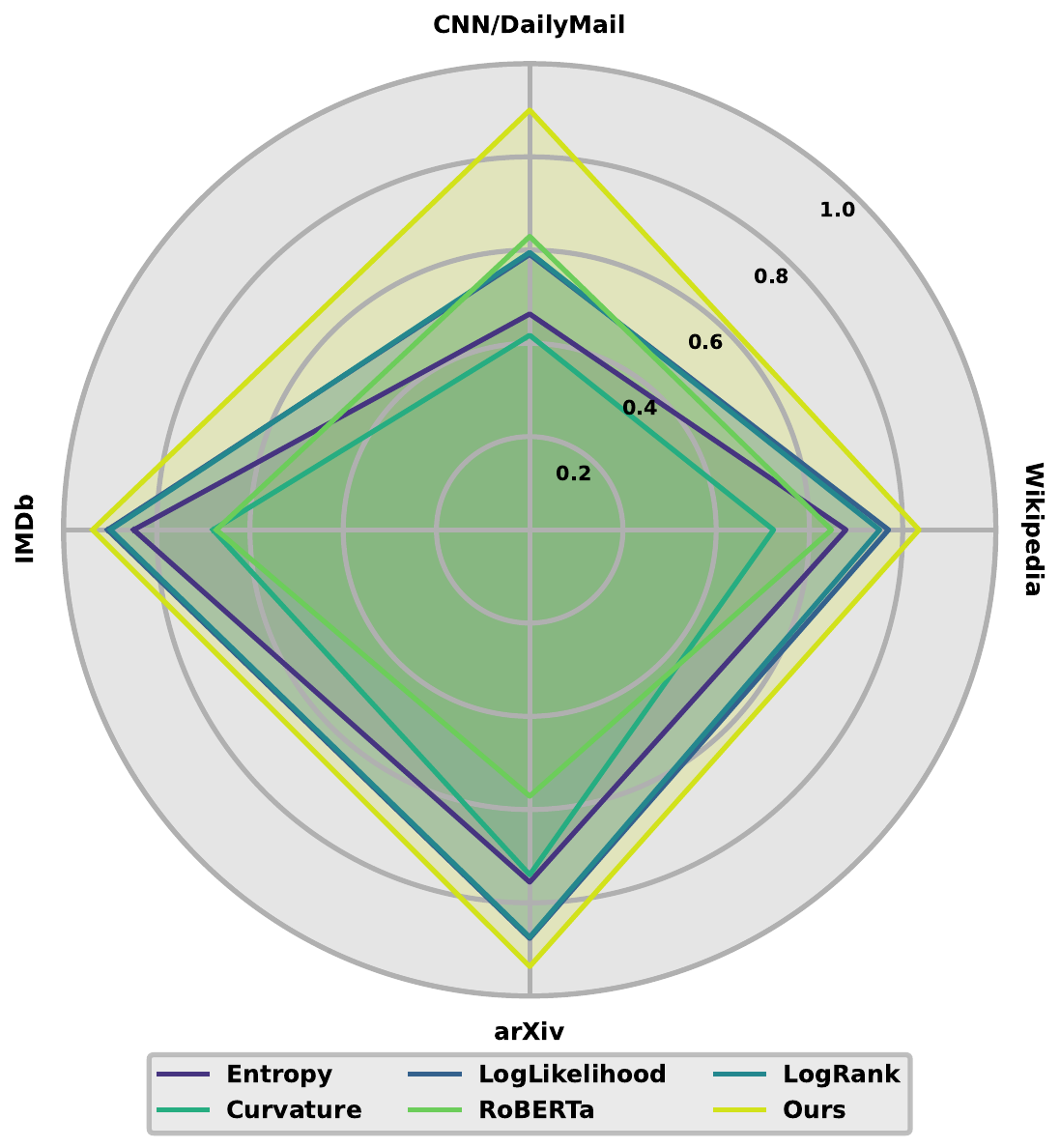}
    }
    \caption{Ternary classification accuracy across models.}
    \label{fig:radar}
\end{figure}

\paragraph*{Baselines}
In our comparative study, the baselines include four zero-shot statistical methods, Entropy~\cite{10.5555/3053718.3053722}, LogLikelihood~\cite{gehrmann-etal-2019-gltr}, LogRank~\cite{su-etal-2023-detectllm}, and Curvature~\cite{pmlr-v202-mitchell23a}, together with a data-driven detector based on the RoBERTa model~\cite{uchendu-etal-2021-turingbench-benchmark}. These methods were originally developed for the binary task of distinguishing human-written from machine-generated text rather than for AI-role attribution. Nevertheless, in the present task setting, the assistive and creative roles differ partly in the degree of machine involvement: assistive writing remains more closely tied to the original human-written text, whereas creative writing is generated more independently by the language model. The baseline methods therefore remain applicable insofar as they capture a notion of machine-likeness.

\paragraph*{Metrics}
Three metrics are used throughout the experiments. Discriminability is measured by the area under the receiver operating characteristic curve (AUC) and the classification accuracy (ACC). AUC assesses discriminability independently of any particular decision threshold, whereas ACC is computed at the threshold selected via cross-validation. Text quality is measured by perplexity, which quantifies the uncertainty of a language model in predicting a given text, with a lower value indicating a better fit.

\paragraph*{Hardware and software}
All experiments were conducted on a server equipped with an NVIDIA A100 GPU. The implementation was based on Python and PyTorch. All language models were publicly available open-source checkpoints from the Hugging Face Hub.

\subsection{Discriminability Analysis}
We analyse the discriminability of the proposed system across different datasets and language models in comparison to the baselines. Our experimentation is designed to examine three complementary aspects: performance on pairwise binary classification, performance on joint ternary classification and the distribution of discriminative features.

\paragraph*{Pairwise analysis} Table~\ref{tab:results} reports the performance on the three pairwise binary classification tasks, namely human versus assistive ($\mathbb{H}$ vs $\mathbb{A}$), human versus creative ($\mathbb{H}$ vs $\mathbb{C}$), assistive versus creative ($\mathbb{A}$ vs $\mathbb{C}$) and average. For each method, the decision threshold is determined automatically via 10-fold cross-validation. In each round, one fold is held out for testing, while the remaining nine folds are used to select the threshold that yields the best performance. The selected threshold is then applied to the held-out fold. This process is repeated for all ten folds, and the results are averaged across folds. The baseline methods each produce a single scalar score, as they were originally designed for the binary task of distinguishing human-written from AI-generated content. By contrast, our method produces multiple features since it was originally designed for multi-class role attribution, in which a separate $p$-value is computed for each AI-related role. To adapt our method to the pairwise analysis, these multiple features are decomposed into distinct decision variables for each binary subtask. Specifically, $p_{\mathbb{A}}$ is used for distinguishing between $\mathbb{H}$ and $\mathbb{A}$, $p_{\mathbb{C}}$ is used for distinguishing between $\mathbb{H}$ and $\mathbb{C}$, $p_{\mathbb{C}} - p_{\mathbb{A}}$ is used for distinguishing between $\mathbb{A}$ and $\mathbb{C}$. Overall, the proposed method consistently achieves the best performance across all datasets and both language models. It attains an average AUC of 0.91 and an average ACC of 0.88 under GPT-2, and the performance further improves to 0.99 and 0.95, respectively, under LLaMA-3-Instruct, an improvement that may partly reflect the model's stronger ability to understand and classify the role. The advantage of the proposed method is particularly pronounced on the more challenging role-attribution task ($\mathbb{A}$ vs $\mathbb{C}$), in contrast to the more conventional task of detecting AI-generated text ($\mathbb{H}$ vs $\mathbb{A}$ and $\mathbb{H}$ vs $\mathbb{C}$). The baseline methods exhibit substantial degradation when both classes correspond to AI-generated content and differ only in the role played by the model. This observation suggests that existing reactive detectors primarily capture the presence of machine generation, but are limited in their ability to distinguish between different modes of generation. These results therefore highlight the need for a role-attribution framework in this context.

\begin{figure*}[!t]
\centering
\includegraphics[width=0.9\linewidth]{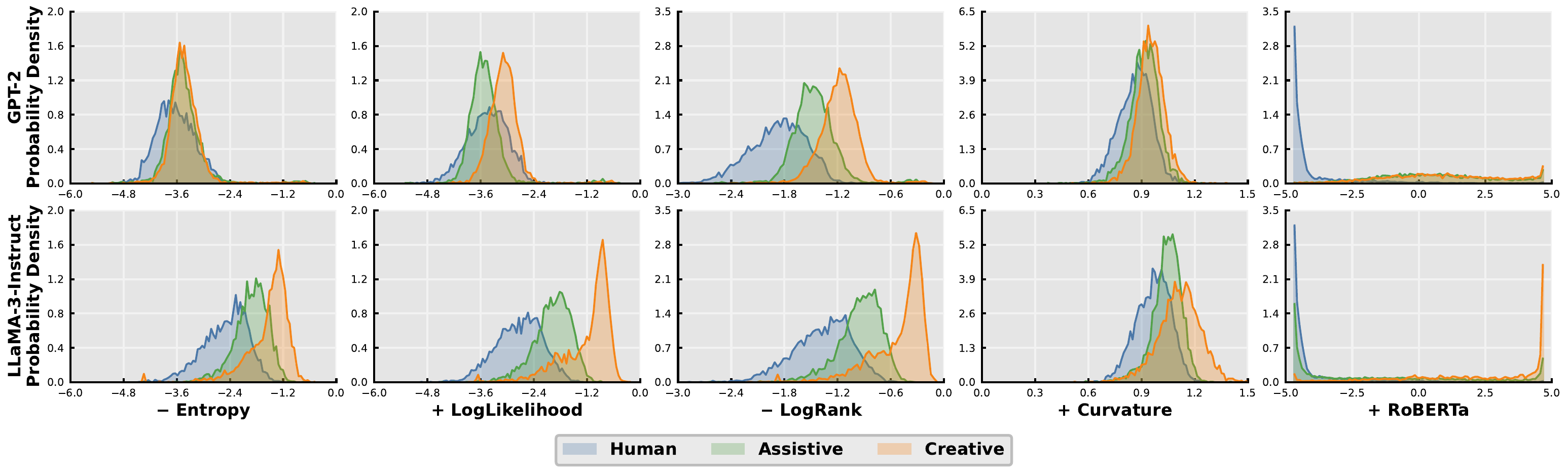}
\caption{Distribution of baseline features across models.}
\label{fig:baseline_dist}
\end{figure*}

\begin{figure}[t!]
    \centering
    \subfloat[GPT-2 (124 million parameters)]{
    	\includegraphics[width=0.45\columnwidth]{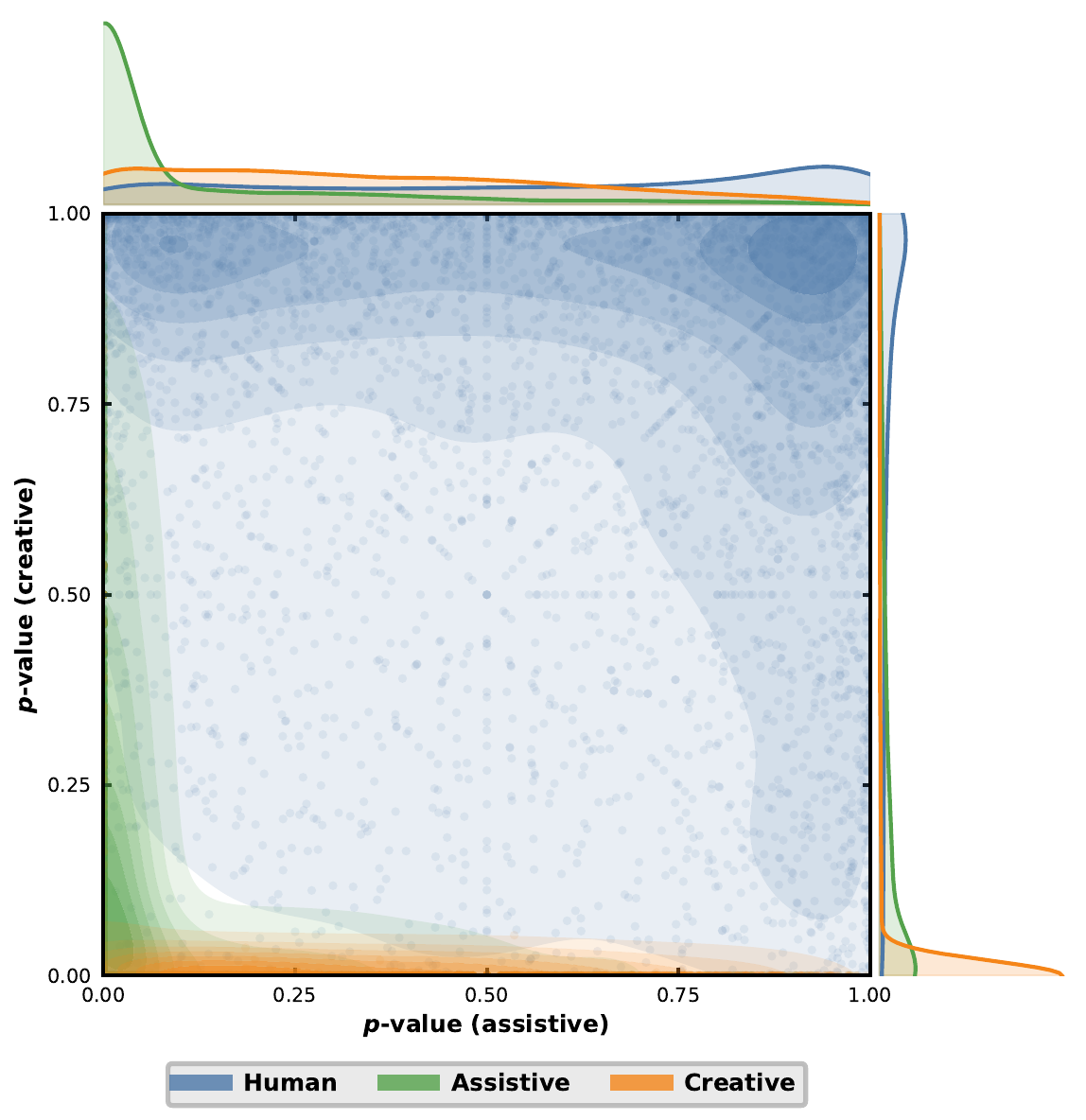}
    }
	\hfill
    \subfloat[LLaMA-3-Instruct (3 billion parameters)]{
        \includegraphics[width=0.45\columnwidth]{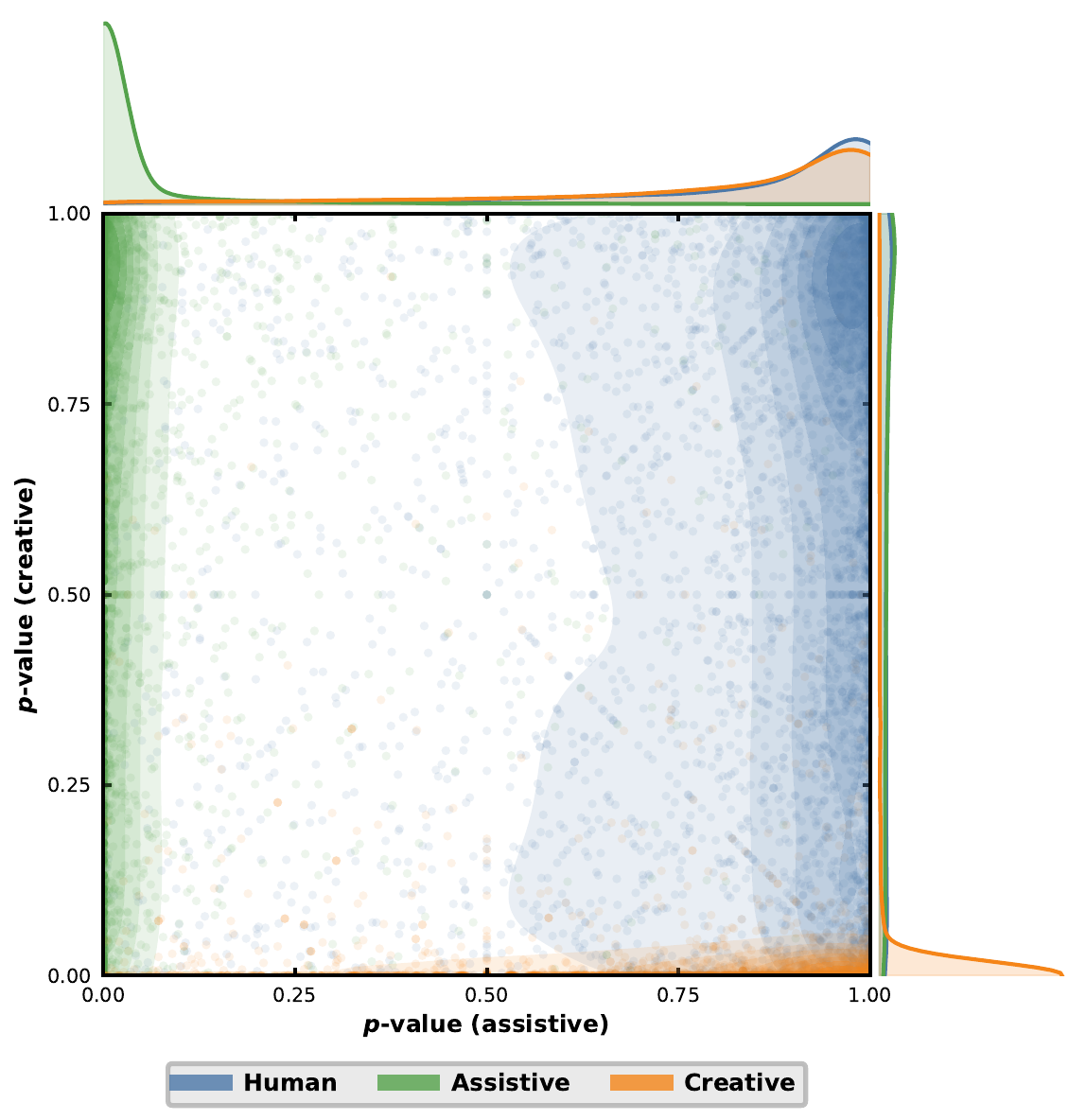}
    }
    \caption{Distribution of role-specific $p$-values across models.}
    \label{fig:ours_dist}
\end{figure}

\paragraph*{Joint analysis} Figure~\ref{fig:radar} evaluates the performance on joint ternary classification. For each method, the threshold selection procedure is likewise performed by 10-fold cross-validation. For each baseline method, the single scalar feature is assumed to be ordered such that samples from class $\mathbb{H}$ tend to yield the smallest values, followed by $\mathbb{A}$, and then $\mathbb{C}$, where larger values indicate stronger evidence of machine generation. The optimal threshold pair is obtained by exhaustively searching the ordered feature axis so as to partition the feature axis into three regions that maximise the three-class accuracy on the nine training folds. In contrast, the proposed method involves only one threshold $\theta$, applied to the $p$-value of each role. If none of the role-specific $p$-values is smaller than $\theta$, the sample is regarded as lacking sufficient statistical significance for any role. In other words, a text is classified as human-written if all $p$-values exceed $\theta$; otherwise, it is assigned to the role with the smallest $p$-value. The optimal threshold $\theta$ is selected by searching over the ordered values of $\min(p_{\mathbb{A}}, p_{\mathbb{C}})$ on the training folds. A similar trend is observed for the ternary classification task, where the proposed method generally attains the highest performance, or performance comparable to the strongest baseline, across the datasets and both language models. The advantage is particularly pronounced on CNN/DailyMail, whose concepts and contents contain the largest number of words. In this dataset, the prompts for creative writing are based on news highlights rather than brief concepts, similar to the prompts used to generate the assistive writing. Consequently, the prompts for both creative and assistive writing contain substantial amounts of human-written text, making both AI-content detection and AI-role attribution more challenging.

\begin{table*}[t]
\centering
\caption{
Evaluation of perplexity across models and datasets
}
\label{tab:pplx}

\resizebox{\linewidth}{!}{%
\begin{tabular}{c  ccccc  ccccc}
\toprule

\multicolumn{1}{c}{} &
\multicolumn{5}{c}{GPT-2} &
\multicolumn{5}{c}{LLaMA-3-Instruct} \\

\cmidrule(lr){2-6}
\cmidrule(lr){7-11}

& \makebox[1.4cm][c]{$\mathbb{H}$}
& \makebox[1.4cm][c]{$\mathbb{A}$ (unbiased)}
& \makebox[1.4cm][c]{$\mathbb{A}$ (biased)}
& \makebox[1.4cm][c]{$\mathbb{C}$ (unbiased)}
& \makebox[1.4cm][c]{$\mathbb{C}$ (biased)} 
& \makebox[1.4cm][c]{$\mathbb{H}$}
& \makebox[1.4cm][c]{$\mathbb{A}$ (unbiased)}
& \makebox[1.4cm][c]{$\mathbb{A}$ (biased)}
& \makebox[1.4cm][c]{$\mathbb{C}$ (unbiased)}
& \makebox[1.4cm][c]{$\mathbb{C}$ (biased)} 
\\

\midrule

\multicolumn{1}{l}{IMDb}
& \multicolumn{1}{c}{47.15} & \multicolumn{1}{c}{34.69} & \multicolumn{1}{c}{46.19} & \multicolumn{1}{c}{21.76} & \multicolumn{1}{c}{27.09}
& \multicolumn{1}{c}{28.02} & \multicolumn{1}{c}{8.47} & \multicolumn{1}{c}{9.19} & \multicolumn{1}{c}{2.93} & \multicolumn{1}{c}{4.69} \\

\midrule

\multicolumn{1}{l}{CNN/DailyMail}
& \multicolumn{1}{c}{23.14} & \multicolumn{1}{c}{36.18} & \multicolumn{1}{c}{46.53} & \multicolumn{1}{c}{23.95} & \multicolumn{1}{c}{32.75}
& \multicolumn{1}{c}{12.84} & \multicolumn{1}{c}{8.39} & \multicolumn{1}{c}{7.70} & \multicolumn{1}{c}{8.21} & \multicolumn{1}{c}{8.89} \\

\midrule

\multicolumn{1}{l}{Wikipedia}
& \multicolumn{1}{c}{26.10} & \multicolumn{1}{c}{33.87} & \multicolumn{1}{c}{44.81} & \multicolumn{1}{c}{21.67} & \multicolumn{1}{c}{32.13}
& \multicolumn{1}{c}{9.71} & \multicolumn{1}{c}{6.41} & \multicolumn{1}{c}{6.04} & \multicolumn{1}{c}{2.56} & \multicolumn{1}{c}{3.36} \\

\midrule

\multicolumn{1}{l}{arXiv}
& \multicolumn{1}{c}{44.09} & \multicolumn{1}{c}{36.07} & \multicolumn{1}{c}{49.29} & \multicolumn{1}{c}{24.11} & \multicolumn{1}{c}{33.50}
& \multicolumn{1}{c}{20.11} & \multicolumn{1}{c}{8.71} & \multicolumn{1}{c}{9.56} & \multicolumn{1}{c}{2.98} & \multicolumn{1}{c}{3.87} \\

\midrule

\multicolumn{1}{l}{Overall} & \multicolumn{1}{c}{35.12} & \multicolumn{1}{c}{35.20} & \multicolumn{1}{c}{46.71} & \multicolumn{1}{c}{22.87} & \multicolumn{1}{c}{31.37} & \multicolumn{1}{c}{17.67} & \multicolumn{1}{c}{7.99} & \multicolumn{1}{c}{8.12} & \multicolumn{1}{c}{4.17} & \multicolumn{1}{c}{5.20} \\

\bottomrule
\end{tabular}
}
\end{table*}

\paragraph*{Distributional analysis} 
Figure~\ref{fig:baseline_dist} and Figure~\ref{fig:ours_dist} visualise the feature distributions across different classes for the baseline methods and the proposed method, respectively. The results from different datasets are aggregated, whereas different language models are shown separately. For the baseline methods, the assistive samples tend to occupy an intermediate region between the human and creative samples along the feature axis. This observation supports the assumption that larger values indicate stronger evidence of machine generation. While the human and creative samples are often reasonably separable, the assistive samples still overlap substantially with both. The data-driven detector yields scores that are highly concentrated near the two extremes, yet still provides only limited separation of the assistive samples. These results suggest that the baseline methods primarily capture a notion of machine-likeness, which is insufficient to distinguish different roles in AI generation. In contrast, the proposed method forms a two-dimensional feature space. Human-written samples are concentrated near the upper-right region, where both $p$-values are large, indicating that neither the assistive nor the creative role is statistically significant. Assistive and creative samples are instead concentrated near the left and lower boundaries, respectively. Compared with the results associated with GPT-2, those associated with LLaMA-3-Instruct exhibit clearer separation between the creative and assistive samples. These observations suggest that the proposed method captures role-specific evidence beyond a one-dimensional notion of machine-likeness, thereby enabling more reliable discrimination between different forms of AI involvement.

\subsection{Perplexity Analysis}
Table~\ref{tab:pplx} reports the perplexity of the human-written texts, the unbiased assistive and creative texts, and their biased counterparts in order to evaluate the impact of biasing the generation distribution on text quality. Perplexity measures the uncertainty of a language model when predicting a given text. A lower perplexity score indicates that the model is less confused and better at predicting the words in a given sample. Although perplexity does not directly correspond to fluency or coherence as perceived by humans, it is often used as a proxy for text quality. We measure self-perplexity, in which the same model is used for both text generation and text evaluation. Note that self-perplexity is model-specific, so these values should be read relative to each model's own baseline rather than compared across models in absolute terms.

Overall, biasing tends to increase self-perplexity, although the effect is not uniform across all datasets. Creative writing often yields the lowest perplexity because it is mostly created by the language model without substantial human-written prompts and thus conforms most closely to the model's internal distribution. By contrast, human-written text generally exhibits higher perplexity because it is not produced by the model and may therefore deviate from its preferred writing patterns. Assistive writing lies between these two extremes, since it is generated by modifying a human-written text and thus retains characteristics of both human-written and machine-generated language.

\subsection{Robustness Analysis}

\begin{figure}[t!]
    \centering
    \subfloat[GPT-2]{
    	\includegraphics[width=0.45\columnwidth]{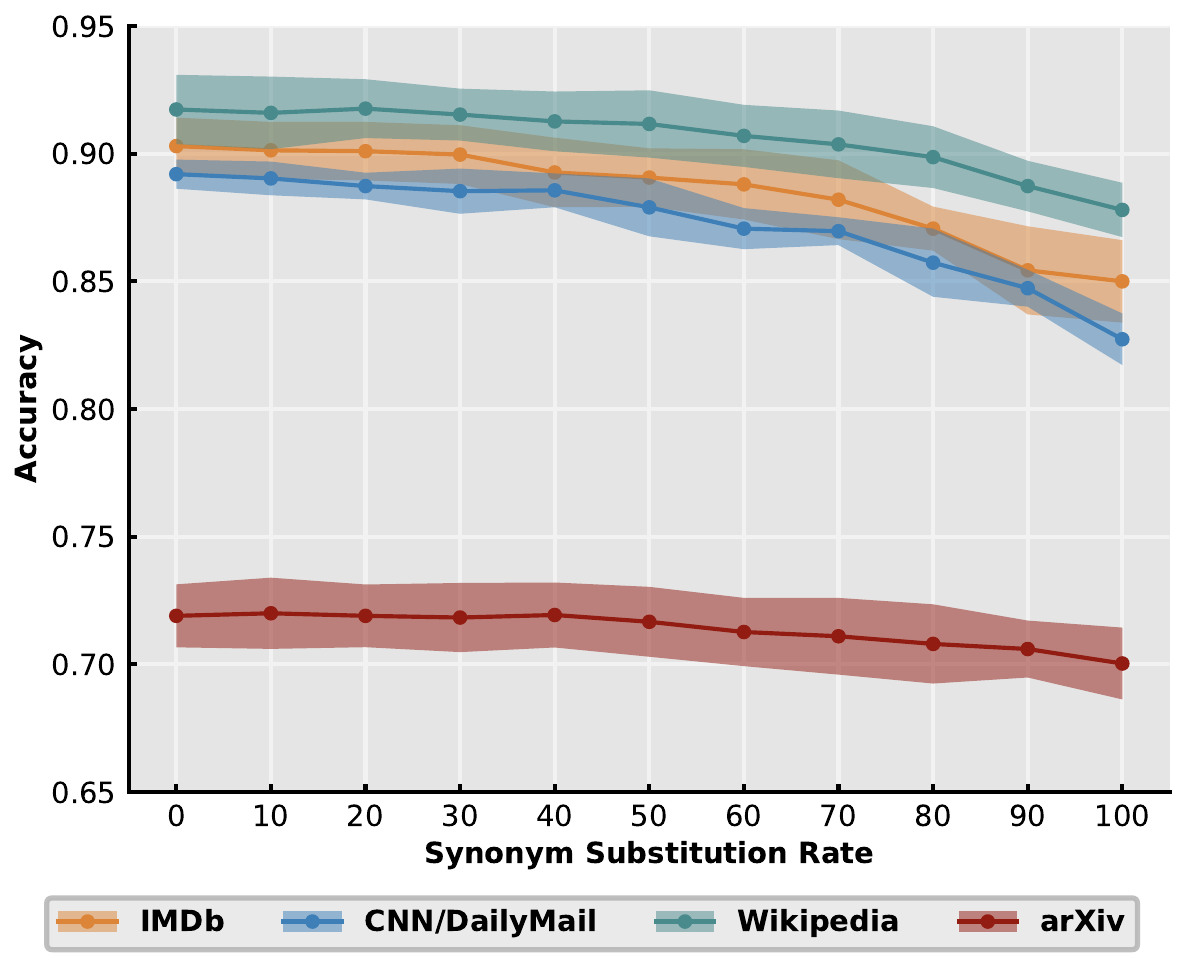}
    }
	\hfill
    \subfloat[LLaMA-3-Instruct]{
        \includegraphics[width=0.45\columnwidth]{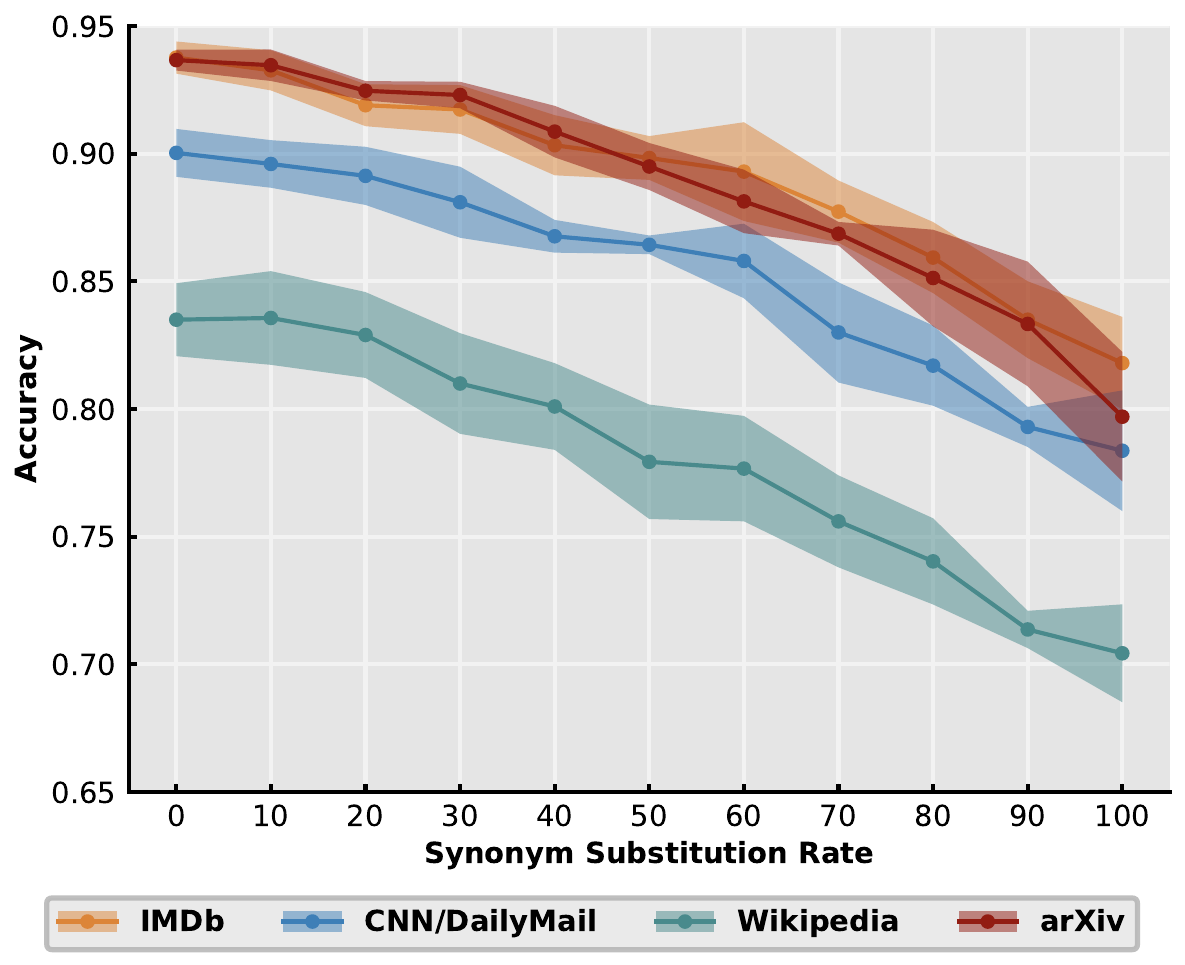}
    }
    \caption{Robustness against synonym substitution across models.}
    \label{fig:robustness_synonym_sub}
\end{figure}

\paragraph*{Synonym Substitution} 
Figure~\ref{fig:robustness_synonym_sub} evaluates the robustness of the proposed method under synonym substitution. In practice, AI-generated text may be humanised to evade or confuse detection. Synonym substitution is adopted because it constitutes one of the simplest and most controllable forms of such post-editing. Compared with stronger forms of paraphrasing, synonym substitution largely preserves the original semantics and sentence structure, thereby providing a standardised perturbation that isolates the effect of lexical variation. In our experiments, a proportion of eligible words in each text is randomly replaced with synonyms obtained from the synsets of WordNet, a large lexical database of English in which words are grouped into synonym sets interlinked through semantic relations. Only content words with a valid part-of-speech tag (nouns, verbs, adjectives and adverbs) are considered, whereas function words and punctuation marks are preserved. The synonym substitution rate denotes the percentage of eligible words that are replaced. As the substitution rate increases, the accuracy decreases gradually for all datasets and both language models. Nevertheless, the degradation remains relatively limited under GPT-2, for which the accuracy decreases by less than approximately 0.06 even when all replaceable words are substituted. Under LLaMA-3-Instruct, the decrease is slightly more pronounced, but still remains below approximately 0.14. Despite this degradation, the proposed method retains reasonably high accuracy over a broad range of substitution rates. In particular, the performance remains above 0.80 for most datasets and above 0.70 for the worst-performing dataset even at high substitution rates. These results suggest that the role information is not carried solely by the content words, but is also distributed across the remaining parts of the text.

\begin{figure}[t!]
    \centering
    \subfloat[GPT-2]{
    	\includegraphics[width=0.45\columnwidth]{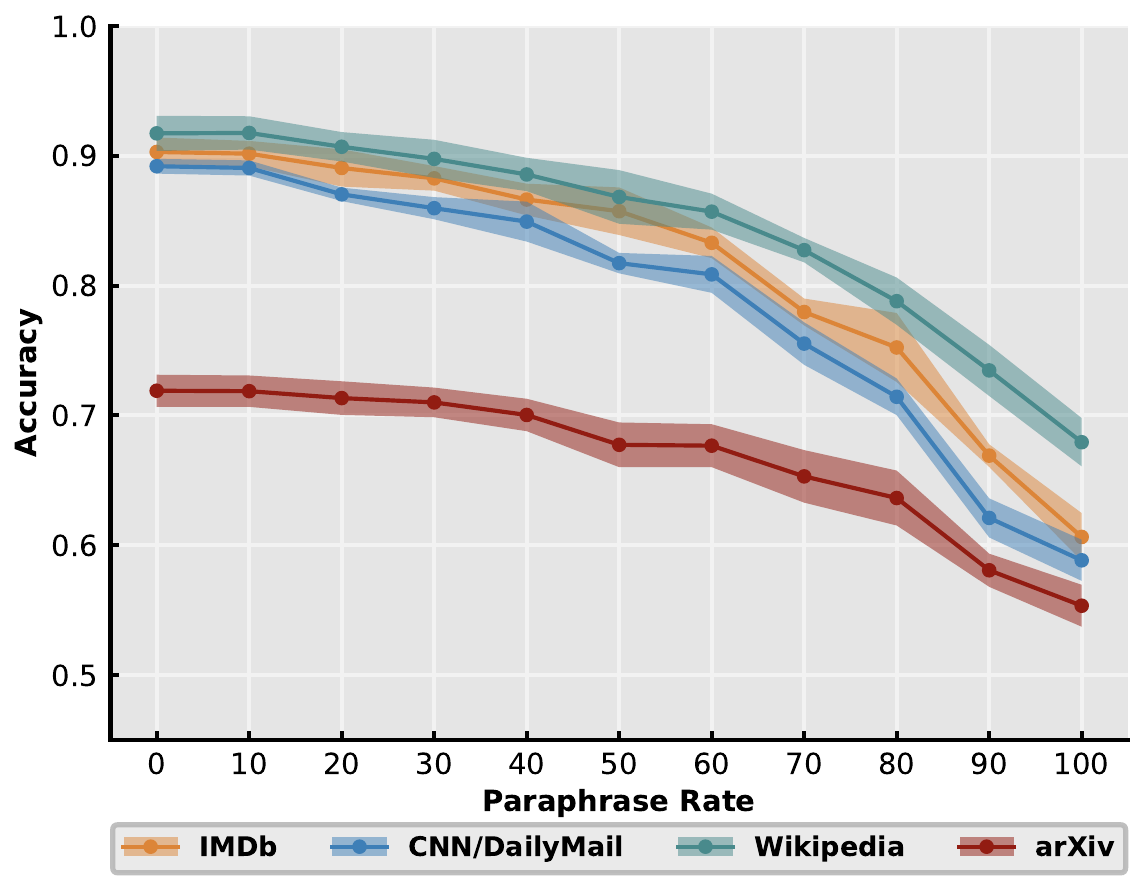}
    }
	\hfill
    \subfloat[LLaMA-3-Instruct]{
        \includegraphics[width=0.45\columnwidth]{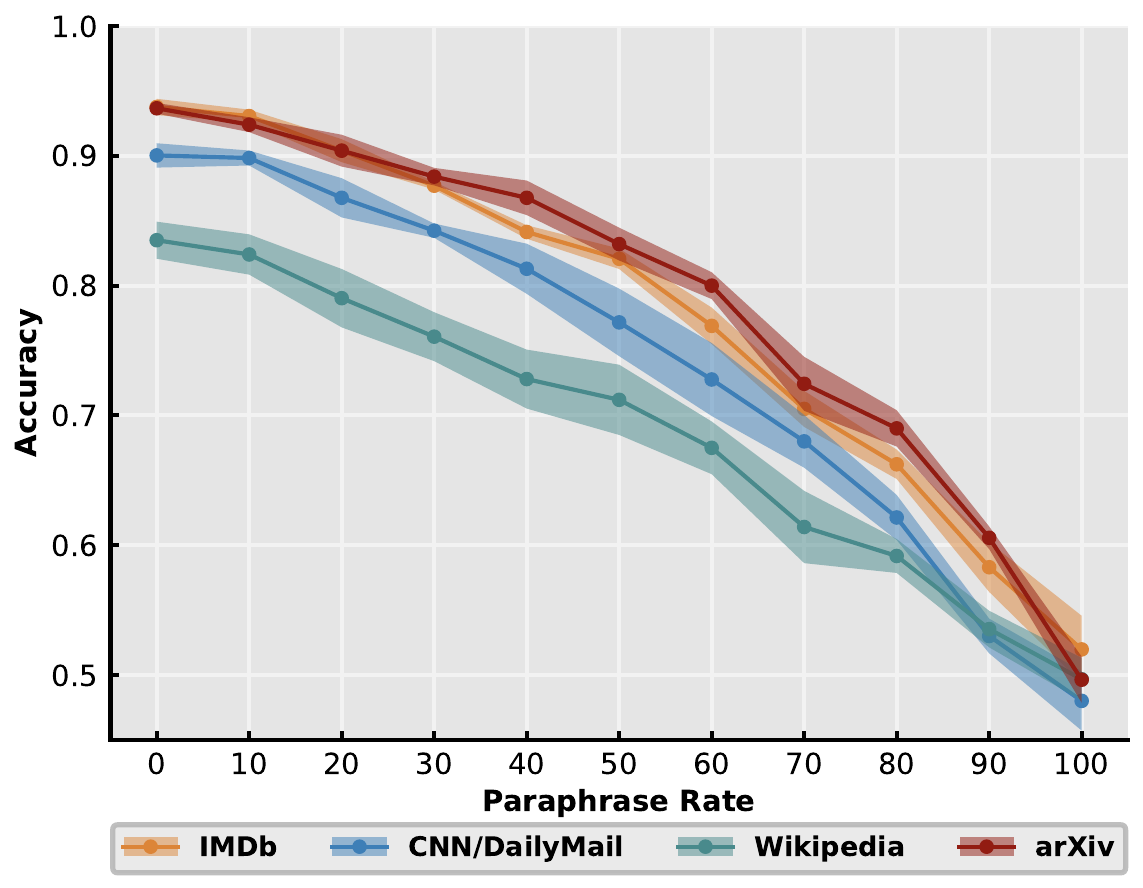}
    }
    \caption{Robustness against sentence paraphrasing across models.}
    \label{fig:robustness_sentence_para}
\end{figure}

\paragraph*{Sentence Paraphrasing}
Synonym substitution perturbs individual words while leaving sentence structure intact. A stronger and more realistic attack is sentence paraphrasing, in which whole sentences are rewritten, altering both wording and structure and thus disturbing a larger portion of the tokens that carry the role signal. To evaluate this, a proportion of the sentences in each text is rewritten by an external paraphrasing model Qwen2.5-1.5B-Instruct, and the paraphrase rate denotes the percentage of sentences rewritten. Figure~\ref{fig:robustness_sentence_para} reports accuracy as the paraphrase rate increases. As expected, paraphrasing degrades accuracy more sharply than synonym substitution, since it disturbs the generated text more extensively. The decline is gradual at low-to-moderate paraphrase rates and becomes more pronounced as most sentences are rewritten.

\begin{figure*}[t!]
    \centering
    \subfloat[GPT-2]{
    	\includegraphics[width=0.99\columnwidth]{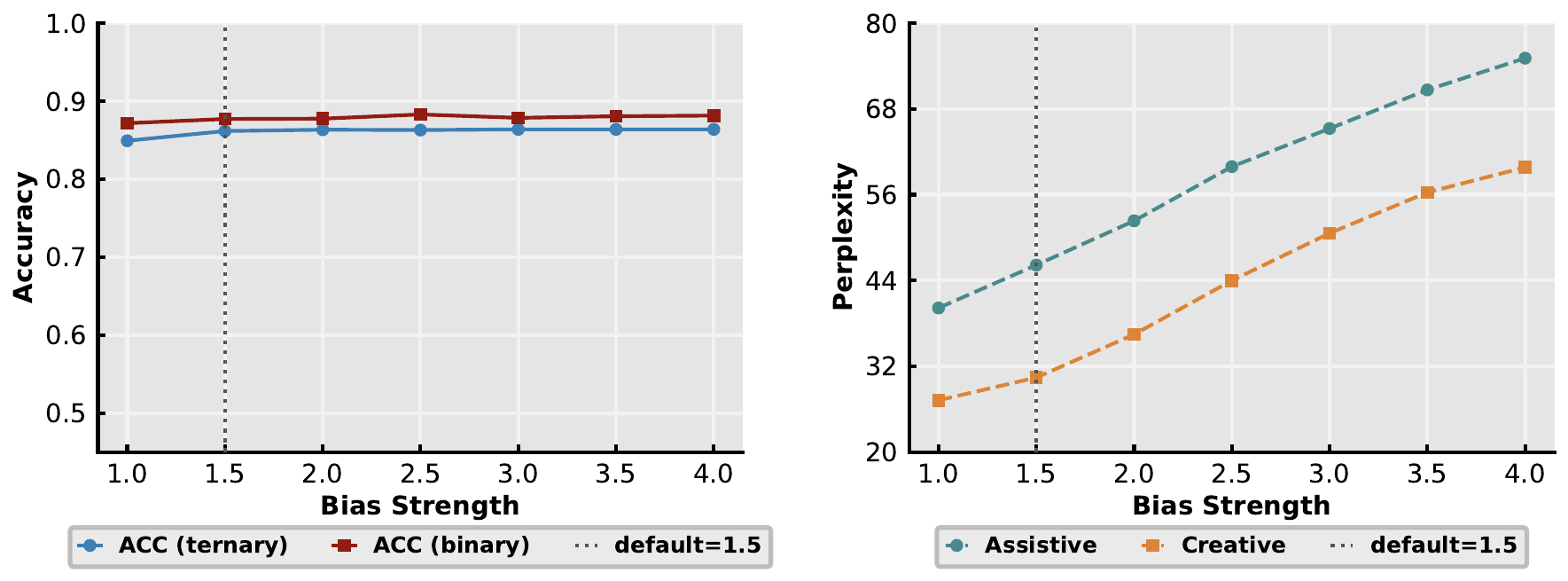}
    }
	\hfill
    \subfloat[LLaMA-3-Instruct]{
        \includegraphics[width=0.99\columnwidth]{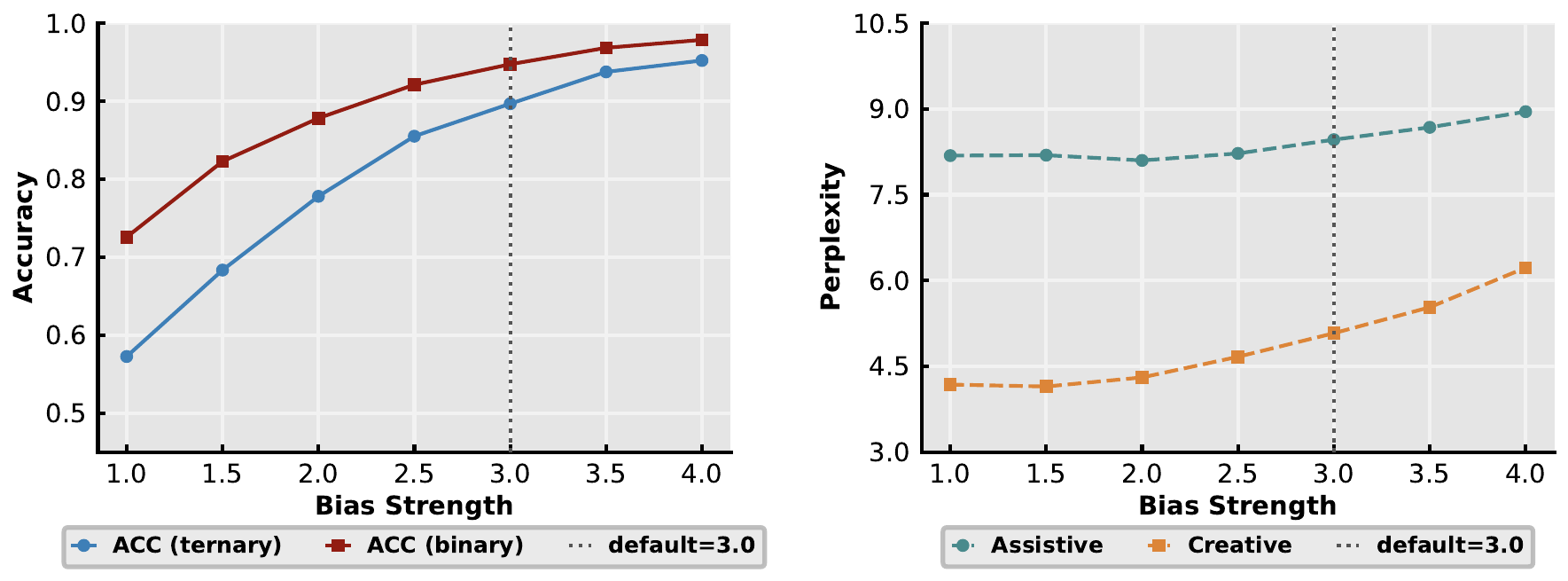}
    }
    \caption{Ablation study on bias strength.}
    \label{fig:bias_strength}
\end{figure*}

\begin{figure*}[t!]
    \centering
    \subfloat[GPT-2]{
    	\includegraphics[width=0.99\columnwidth]{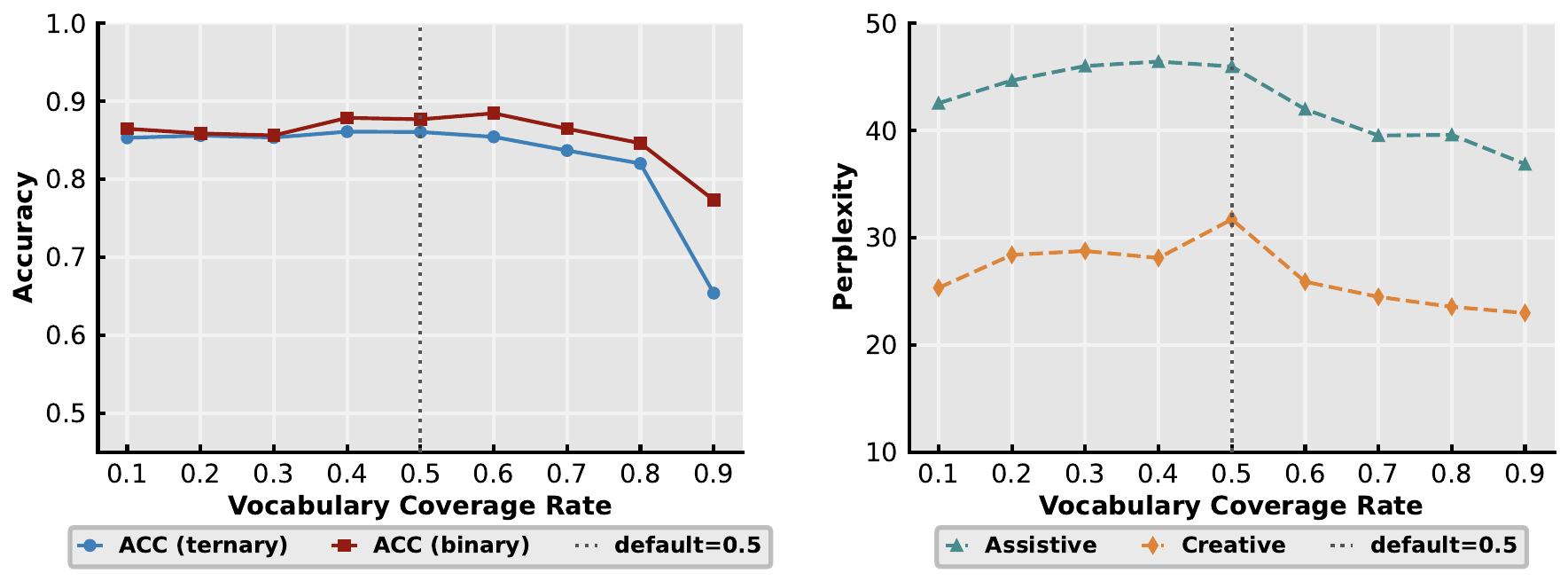}
    }
	\hfill
    \subfloat[LLaMA-3-Instruct]{
        \includegraphics[width=0.99\columnwidth]{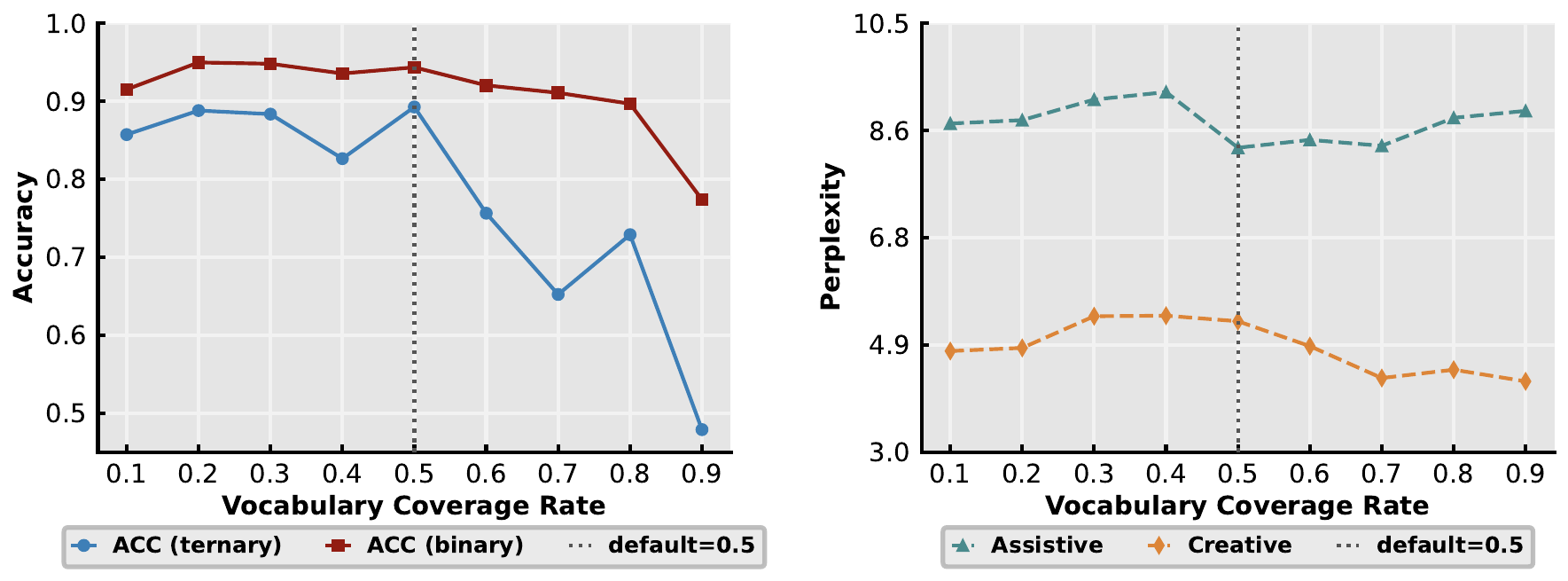}
    }
    \caption{Ablation study on vocabulary coverage rate.}
    \label{fig:vocab_coverage_rate}
\end{figure*}

\section{Ablation}\label{sec:abl}
The proposed system involves four principal hyperparameters: the bias strength $\delta$, the vocabulary coverage rate $q$, the significance threshold $\theta$, and the size of the role space $|\mathcal{R}|$. We examine each in turn, holding the others fixed at their default values.

\subsection{Bias Strength}
The bias strength $\delta$ governs the magnitude of the logit perturbation applied to the role-specific token subset. A larger $\delta$ yields a stronger statistical trace at the cost of a greater departure from the model's natural distribution. Figure~\ref{fig:bias_strength} reports classification accuracy and self-perplexity across a range of $\delta$ values, with the default values used in the other experiments marked for reference. For GPT-2, accuracy reaches a plateau by $\delta = 1.5$ and improves negligibly thereafter, while perplexity rises steeply. For LLaMA-3-Instruct, increasing $\delta$ leads to a substantial improvement in accuracy, while the corresponding increase in perplexity is comparatively modest.

\subsection{Vocabulary Coverage Rate}
The vocabulary coverage rate $q$ sets the size of each role-specific token subset, $\|\mathcal{W}_r\| = q\|\mathcal{V}\|$. Figure~\ref{fig:vocab_coverage_rate} reports classification accuracy and self-perplexity across a range of $q$ values, with the default values used in the other experiments marked for reference. The choice of $q$ balances two opposing tendencies. A smaller subset makes each biased token more distinctive, since a token falling into it by chance is rarer. However, it may also leave fewer tokens to carry the signal, as a biased token is likely to be sampled only if its original probability is already high enough to remain competitive. Accuracy is generally high at smaller rates but becomes increasingly unstable and eventually declines as the rate approaches one. The sharp accuracy decline at high $q$ occurs because role-specific token subsets, drawn independently, overlap heavily as $q$ grows, until the roles can no longer be told apart. Perplexity likewise depends on how much the bias distorts sampling. When $q$ is small, the bias affects too few tokens to change the sampling distribution much; when $q$ is large, it lifts almost every token by a similar amount and so again leaves the distribution largely unchanged.

\begin{table}[t]
\caption{Ablation study on significance threshold.}
\label{tab:theta_sensitivity}
\centering
\resizebox{\linewidth}{!}{%
\begin{tabular}{lccc@{\hskip 10pt}ccc}
\toprule
 & \multicolumn{3}{c}{GPT-2} & \multicolumn{3}{c}{LLaMA-3-Instruct} \\
\cmidrule(lr){2-4}\cmidrule(lr){5-7}
 & Acc$_{\mathrm{specific}}$ & Acc$_{\mathrm{agnostic}}$ & Absolute Difference & Acc$_{\mathrm{specific}}$ & Acc$_{\mathrm{agnostic}}$ & Absolute Difference \\
\midrule
IMDb & 0.903 & 0.903 & 0.000 & 0.938 & 0.938 & 0.000 \\
CNN/DailyMail & 0.892 & 0.886 & 0.006 & 0.901 & 0.902 & 0.001 \\
Wikipedia & 0.917 & 0.919 & 0.002 & 0.834 & 0.838 & 0.004 \\
arXiv & 0.719 & 0.720 & 0.001 & 0.937 & 0.933 & 0.004 \\
\midrule
\textbf{Overall} & \textbf{0.858} & \textbf{0.857} & \textbf{0.001} & \textbf{0.903} & \textbf{0.903} & \textbf{0.000} \\
\bottomrule
\end{tabular}%
}
\end{table}

\subsection{Significance Threshold}
The significance threshold $\theta$ governs the human-versus-role decision, whereby a text is attributed to a role only when its smallest $p$-value falls below $\theta$. Table~\ref{tab:theta_sensitivity} examines the sensitivity to threshold selection by comparing, for each dataset, ternary accuracy under a dataset-specific threshold selected by cross-validation and a dataset-agnostic threshold estimated from the remaining datasets in a leave-one-dataset-out manner. It is observed that using the dataset-agnostic threshold results in only negligible changes in accuracy on any individual dataset.

\begin{table}[t]
\caption{Ablation study on role-space scalability.}
\label{tab:role_capacity}
\centering
\resizebox{\linewidth}{!}{%
\begin{tabular}{lccccc}
\toprule
Number of Roles $|\mathcal{R}|$ & 2 & 3 & 4 & 5 & 6 \\
Random Chance $1/(|\mathcal{R}|{+}1)$ & 0.333 & 0.250 & 0.200 & 0.167 & 0.143 \\
\midrule
GPT-2 & 0.996 & 0.982 & 0.976 & 0.978 & 0.980 \\
LLaMA-3-Instruct & 0.993 & 0.995 & 0.991 & 0.993 & 0.990 \\
\bottomrule
\end{tabular}%
}
\end{table}

\subsection{Role Space}
The preceding experiments consider a two-role space, and we examine how the method scales as the role space $|\mathcal{R}|$ grows. Table~\ref{tab:role_capacity} reports the result as the role space increases from 2 to 6. To isolate the effect of role count from role semantics, roles are not assigned semantic definitions or inferred through role classification. Instead, each role is treated as an abstract class, identified by the index of its associated vocabulary subset. A fixed set of instructions is used to independently generate text for each role. A text is either human-written or produced under one of the roles, giving $|\mathcal{R}| + 1$ possible classes in total. Random guessing therefore gives an accuracy of $1/(|\mathcal{R}|+1)$. For both models, the accuracy remains near-perfect throughout, staying above $0.97$ even as the chance level under random guessing falls. Within the examined range of role-space sizes, the encoding and decoding mechanism scales with negligible loss, suggesting that errors in the full pipeline are more likely to arise from role classification at the prompt-analysis stage than from role encoding and decoding.

\section{Conclusion}\label{sec:con}
This study has investigated the problem of tracing the role played by AI in human-machine collaboration, moving beyond a binary classification between human-written and machine-generated content. A methodology has been proposed to infer the latent role specified by the prompt, encode this role into the generation process and subsequently decode it from the resulting text. In this manner, the functional role of AI can remain observable even after the original prompt and dialogue context are no longer available. The experimentation focuses on a representative scenario in which AI acts either as an assistive or a creative agent, yet the implications of this study extend beyond these particular forms of participation. The experimental results demonstrate that the proposed method achieves effective discrimination between assistive and creative roles, remains reasonably robust under textual perturbations and preserves acceptable text quality despite the introduction of lexical bias during language generation. This study reflects the view that the significance of AI lies not merely in whether it has participated, but in how it has participated.

Several directions remain for future research. First, a more intricate form of collaboration may involve a substantially richer taxonomy of roles, together with hierarchical relationships between them. In addition, an extension beyond natural language to other modalities, including visual and auditory content, is expected. Furthermore, tracing roles over multi-turn interaction merits further investigation, since the role of AI may evolve dynamically according to the conversational context. We envision that this study may contribute towards a deeper understanding of human-machine symbiosis, in which the role assumed by AI becomes more transparent and traceable.

\newpage
\section*{Acknowledgements}
This work was supported in part by the Japan Society for the Promotion of Science (JSPS) under KAKENHI Grants JP21H04907 and JP24H00732, and in part by the Japan Science and Technology Agency (JST) under the CREST Grants JPMJCR20D3 and JPMJCR2562, including the AIP Challenge Program, and under the AIP Acceleration Grant JPMJCR24U3 and under the K Program Grant JPMJKP24C2.

\bibliography{Transactions-Bibliography/bstcontrol, Bib/bib_role}
\bibliographystyle{Transactions-Bibliography/IEEEtran}

\vspace{11em}
\begin{IEEEbiographynophoto}{Ching-Chun Chang} (Senior Member, IEEE) received the PhD in Computer Science from the University of Warwick, UK, in 2019. He is currently affiliated with the National Institute of Informatics, Japan, as a Project Assistant Professor. He also serves as a Visiting Researcher at Peking University, China, and a Distinguished Professor at Hangzhou Dianzi University, China. He is an IEEE Senior Member, an IET Member, and an IEICE Member, and has been elected as a Distinguished Scientist of the International Academy of Artificial Intelligence Sciences for his contributions to trustworthy artificial intelligence. He participated in the Short-Term Scientific Mission supported by European Cooperation in Science and Technology Actions at the Faculty of Computer Science, Otto von Guericke University of Magdeburg, Germany, in 2016. He was granted the Marie-Curie Fellowship and participated in the Research and Innovation Staff Exchange supported by Marie Skłodowska-Curie Actions at the Department of Electrical and Computer Engineering, New Jersey Institute of Technology, USA, in 2017. He was a Visiting Scholar at the School of Computing and Mathematics, Charles Sturt University, Australia, in 2018, and at the School of Information Technology, Deakin University, Australia, in 2019. He was a Research Fellow at the Department of Electronic Engineering, Tsinghua University, China, in 2020. His research interests include artificial intelligence, biometrics, cryptography, cybersecurity, evolutionary computation, forensics, steganography, and watermarking.
\end{IEEEbiographynophoto}

\newpage
\begin{IEEEbiographynophoto}{Yuchen Guo} received the BS degree from the Faculty of Informatics, Beijing University of Technology in 2022 and the MS degree from the University of Tokyo in 2025. He is currently pursuing the PhD degree at the University of Tokyo, while also working as a Research Assistant at the National Institute of Informatics, Japan. His research focuses on AI security and NLP security, with particular interests in the security, robustness, and trustworthiness of intelligent systems.
\end{IEEEbiographynophoto}

\begin{IEEEbiographynophoto}{Hanrui Wang}
received the BS degree in Electronic Information Engineering from Northeastern University (China) in 2011. After working in the IT industry and rising to a director-level position, he transitioned to academic research in 2019 and earned the PhD in Computer Science from Monash University, Australia, in January 2024. He is currently an Assistant Professor at the National Institute of Informatics, Japan. His research focuses on AI security and privacy, with emphasis on adversarial machine learning, model inversion, and trustworthy multimedia systems. He has published multiple papers in leading journals such as IEEE TIFS, IEEE TDSC, ACM TOMM, and J-STARS, as well as in reputable international conferences including WWW, WACV, FG, ICPR, ICASSP, and ICMI.
\end{IEEEbiographynophoto}

\begin{IEEEbiographynophoto}{Timo Spinde} received the PhD in Computer Science from the University of Göttingen, Germany, in 2024, graduating summa cum laude. He also holds a master’s degree in Social and Economic Data Analysis from the University of Konstanz and completed dual bachelor’s degrees in Computer Science and Media and Communication at the University of Passau. He is currently a Postdoctoral Researcher with the National Institute of Informatics, Tokyo, Japan, and the University of Göttingen, Germany. He is also the initiator and coordinator of the Media Bias Group, an international research network dedicated to interdisciplinary research on media bias, and has founded multiple startups. His research interests include media bias detection, natural language processing, artificial intelligence, computational social science, and information quality.\end{IEEEbiographynophoto}

\begin{IEEEbiographynophoto}{Isao Echizen} (Senior Member, IEEE) received BS, MS, and DE degrees from the Tokyo Institute of Technology, Japan, in 1995, 1997 and 2003, respectively. He joined Hitachi, Ltd. in 1997 and until 2007 was a Research Engineer in the company's systems development laboratory. He is currently a Director and Professor of the Information and Society Research Division, as well as a Director of the Global Research Center for Synthetic Media, at the National Institute of Informatics; a Professor in the Department of Information and Communication Engineering, Graduate School of Information Science and Technology, the University of Tokyo; and a Professor in the Graduate University for Advanced Studies (SOKENDAI), Japan. He was a Visiting Professor at the Tsuda University, Japan; at the University of Freiburg, Germany; and at the University of Halle-Wittenberg, Germany. He is currently engaged in research on AI security, multimedia security and multimedia forensics, serving as a Research Director for the CREST FakeMedia project and the K Program SYNTHETIQ X project of the Japan Science and Technology Agency (JST). He received the Commendation for Science and Technology by the Minister of Education, Culture, Sports, Science and Technology (Research Category) in 2025. He also received the IEICE Best Paper Award in 2023; the IPSJ Best Paper Awards in 2005 and 2014; the IPSJ Nagao Special Researcher Award in 2011; the DOCOMO Mobile Science Award in 2014; the IISEC Information Security Cultural Award in 2016; and the IEEE WIFS Best Paper Award in 2017. He is an IEICE Fellow, an IPSJ Fellow, an IEEE Senior Member, an IFIP Japanese Representative, and an APSIPA Vice President.
\end{IEEEbiographynophoto}

\end{document}